\documentclass[letterpaper]{article} 
\usepackage{aaai2027}  
\usepackage[hyphens]{url}  
\usepackage{graphicx} 
\usepackage{natbib}  
\usepackage{caption} 
\usepackage{algorithm}
\usepackage{algorithmic}

\usepackage{newfloat}
\usepackage{listings}
\DeclareCaptionStyle{ruled}{labelfont=normalfont,labelsep=colon,strut=off} 
\floatstyle{ruled}
\newfloat{listing}{tb}{lst}{}
\floatname{listing}{Listing}

\usepackage{booktabs}

\usepackage{soul}
\usepackage{url}

\usepackage{amsfonts}
\usepackage{textcomp}
\usepackage[utf8]{inputenc}
\usepackage{units}
\usepackage{multirow}
\usepackage{amsmath}
\usepackage{amsthm}
\usepackage{graphicx}

\title{Study and Improvement of Search Algorithms in Stochastic Perfect-Information Games}
\author{
    Quentin Cohen-Solal\corresponding, Tristan Cazenave
}
\affiliations{
LAMSADE, Université@ Paris-Dauphine, PSL, CNRS, Paris, France\\
quentin.cohen-solal@dauphine.psl.eu
}

\begin{document}
\maketitle
\begin{abstract}
In this article, we generalize Unbounded Minimax, the state-of-the-art
search algorithm for deterministic zero-sum two-player games with
perfect information to the framework of stochastic zero-sum two-player
games with perfect information. We experimentally show that this generalized
algorithm achieves better performance than the main stochastic search
algorithms.
\end{abstract}



\global\long\def\minibaln{\mathrm{Minibal_{n}}}%
\global\long\def\minibalp{\mathrm{Minibal_{+}}}%
\global\long\def\minibalc{\mathrm{Minibal_{c}}}%
\global\long\def\indicatrice#1#2{\boldsymbol{1}_{#1}\left(#2\right)}%

\global\long\def\et{\ \wedge\ }%
\global\long\def\terminal#1{\mathrm{terminal}\left(#1\right)}%
\global\long\def\joueur{\mathrm{j}}%
\global\long\def\joueurUn{\mathrm{1}}%
\global\long\def\joueurDeux{\mathrm{2}}%
\global\long\def\fbin{\mathrm{f_{b}}}%
\global\long\def\Actions{\mathcal{A}}%
\global\long\def\actions#1{\mathrm{actions}\left(#1\right)}%
\global\long\def\etats{\mathcal{S}}%
\global\long\def\random{\mathrm{random}\left(\right)}%
\global\long\def\premier#1{\mathrm{first_{\text{–}}player}\left(#1\right)}%
\global\long\def\balanced#1{\mathrm{balanced_{\text{–}}player}\left(#1\right)}%
\global\long\def\rootplayer#1{\mathrm{root_{\text{–}}player}\left(#1\right)}%
\global\long\def\terminalrandom{\mathrm{t_{r}}}%
\global\long\def\hfb{\mathrm{b_{t}}}%
\global\long\def\hfp{\mathrm{p_{t}}}%
\global\long\def\hadapt{f_{\theta}}%
\global\long\def\hadaptnum#1{f_{\theta_{#1}}}%
\global\long\def\itreeset{\mathbf{T_{i}}}%
\global\long\def\treeset{\mathbf{T}}%
\global\long\def\rootset{\mathbf{R}}%
\global\long\def\hterminal{f_{\mathrm{t}}}%
\global\long\def\rnap{\mathrm{p_{rna}}}%
\global\long\def\rnab{\mathrm{b_{rna}}}%
\global\long\def\ubfm{\mathrm{UBFM}}%
\global\long\def\ubfmt{\ubfm_{\mathrm{s}}}%
\global\long\def\argmax{\operatorname{\mathrm{arg\,max}}}%
\global\long\def\argmin{\operatorname{\mathrm{arg\,min}}}%
\global\long\def\liste#1#2{\left\{  #1\,|\,#2\right\}  }%
\global\long\def\fterminal{\hterminal}%
\global\long\def\fadapt{\hadapt}%
\global\long\def\id{\mathrm{ID}}%
\global\long\def\minimum{\operatorname*{\mathrm{min}}}%
\global\long\def\Argmax#1{\operatorname{\mathrm{arg\,max}}_{#1}}%
\global\long\def\Argmin#1{\operatorname{\mathrm{arg\,min}}_{#1}}%
\global\long\def\tmax{t_{\mathrm{max}}}%
\global\long\def\negation{\mathop{\neg}}%
\global\long\def\Max#1#2{{\displaystyle \mathop{\mathrm{max}}_{#2}\left(#1\right)}}%
\global\long\def\Min#1#2{{\displaystyle \mathop{\mathrm{min}}_{#2}\left(#1\right)}}%
\global\long\def\rollout#1{\mathrm{rollout}\left(#1\right)}%
\global\long\def\et{\ \wedge\ }%
\global\long\def\ou{\ \vee\ }%
\global\long\def\alphabeta{\alpha\beta}%
\global\long\def\pvs#1{\mathrm{PVS}^{#1}}%
\global\long\def\mtdf#1{\mathrm{MTD(f)}^{#1}}%
\global\long\def\mctsh#1{\mathrm{MCTS}_{\mathrm{h}}^{#1}}%
\global\long\def\mcts#1{\mathrm{MCTS}^{#1}}%
\global\long\def\mctsip#1#2{\mathrm{MCTS}_{\gamma=#2}^{#1}}%
\global\long\def\mctsipm#1{\mathrm{MCTS}_{\alpha\beta}^{#1}}%
\global\long\def\mctsic#1{\mathrm{MCTS}_{\mathrm{max}}^{#1}}%
\global\long\def\joueur{\mathrm{j}}%
\global\long\def\joueurUn{\mathrm{1}}%
\global\long\def\joueurDeux{\mathrm{2}}%
\global\long\def\fbin{\mathrm{f_{b}}}%
\global\long\def\Actions#1{\mathrm{actions}\left(#1\right)}%
\global\long\def\zip{\mathcal{\mathrm{zip}}}%
\global\long\def\etats{\mathcal{S}}%
\global\long\def\mv#1#2{v_{#1,#2}}%
\global\long\def\smv#1{v_{#1}}%
\global\long\def\bv#1#2{v_{#1,#2}}%
\global\long\def\select#1#2{n_{#1,#2}}%
\global\long\def\sselect#1{n_{#1}}%
\global\long\def\sign#1#2{\overline{#1}^{#2}}%
\global\long\def\vsign#1#2{\overline{\mv{#1}{#2}}}%
\global\long\def\bvsign#1#2{\ddddot{\bv{#1}{#2}}}%
\global\long\def\vinf#1#2{v_{#1,#2}^{-}}%
\global\long\def\vinfsign#1#2{\overline{\vinf{#1}{#2}}}%
\global\long\def\vsup#1#2{v_{#1,#2}^{+}}%
\global\long\def\vsupsign#1#2{\overline{\vsup{#1}{#2}}}%
\global\long\def\vsinf#1{v_{#1}^{-}}%
\global\long\def\vssup#1{v_{#1}^{+}}%
\global\long\def\vresol#1#2{c_{#1,#2}}%
\global\long\def\svresol#1{c_{#1}}%
\global\long\def\resol#1#2{r_{#1,#2}}%
\global\long\def\sresol#1{r_{#1}}%
\global\long\def\vresolsign#1#2{\overline{\vresol{#1}{#2}}}%
\global\long\def\svresolsign#1{\overline{\svresol{#1}}}%
\global\long\def\mtdflower{f_{-}}%
\global\long\def\mtdfupper{f_{+}}%
\global\long\def\terminalrandom{\mathrm{t_{r}}}%
\global\long\def\hfb{\mathrm{b_{t}}}%
\global\long\def\hfp{\mathrm{p_{t}}}%
\global\long\def\hadapt{f_{\theta}}%
\global\long\def\hadaptnum#1{f_{\theta_{#1}}}%
\global\long\def\itreeset{\mathbf{T_{i}}}%
\global\long\def\treeset{\mathbf{T}}%
\global\long\def\rootset{\mathbf{R}}%
\global\long\def\hterminal{f_{\mathrm{t}}}%
\global\long\def\rnap{\mathrm{p_{rna}}}%
\global\long\def\rnab{\mathrm{b_{rna}}}%
\global\long\def\ubfm{\mathrm{UBFM}}%
\global\long\def\ubfmt{\ubfm_{\mathrm{s}}}%
\global\long\def\ubfmref{\mathrm{UBFM}_{\mathrm{ref}}}%
\global\long\def\ubfmc{\mathrm{UBFM}_{\mathrm{c}}}%
\global\long\def\ubfms{\mathrm{UBFM}_{\mathrm{s}}}%
\global\long\def\ubfmcs{\mathrm{UBFM}_{\mathrm{c-s}}}%
\global\long\def\liste#1#2{\left\{  #1\,|\,#2\right\}  }%
\global\long\def\fterminal{\hterminal}%
\global\long\def\fadapt{\hadapt}%
\global\long\def\id{\mathrm{ID}}%
\global\long\def\pdiscret{\delta}%
\global\long\def\ended#1{\mathrm{ended}\left(#1\right)}%
\global\long\def\corum{\mathrm{CORUM}}%
\global\long\def\corsum{\mathrm{CORSUM}}%
\global\long\def\actions#1{\mathrm{actions}\left(#1\right)}%

\global\long\def\terminal#1{\mathrm{ended}\left(#1\right)}%

\global\long\def\player#1{\mathrm{player}\left(#1\right)}%
\global\long\def\trunc#1{\mathrm{trunc}\left(#1\right)}%

\global\long\def\time{\mathrm{time}\left(\right)}%

\global\long\def\premier#1{\mathrm{first\ player}\left(#1\right)}%
\global\long\def\orsum{\mathrm{ORSUM}}%
\global\long\def\maxn{\mathrm{Max}^{n}}%
\global\long\def\brs{\mathrm{BRS}}%
\global\long\def\brsp{\mathrm{BRS}^{+}}%

\global\long\def\time{\mathrm{time}\left(\right)}%
\global\long\def\mvs#1{v_{#1}}%
\global\long\def\mctshd#1{\mathrm{MCTS}_{\mathrm{h}}^{#1}{}'}%

\section{Introduction}

Adversarial search in perfect-information games is a core domain of
artificial intelligence, underpinning major theoretical and practical
progress. In the case of two-player zero-sum games, algorithms such
as Minimax and its many extensions provide a well-established framework
for efficiently exploring large combinatorial state spaces. Among
these approaches, Unbounded Minimax~\cite{korf1996best,JMLR:v27:25-2259}
has recently been identified as a state-of-the-art method~\cite{cohen2025study}. 

Unlike traditional depth-limited minimax procedures, Unbounded Minimax
conducts a best-first exploration of the most promising action sequences
without imposing a predefined search horizon. Together with its safe
decision variant~\cite{cohen2025study}, it constitutes the central search component of the
Athénan architecture~\cite{JMLR:v27:25-2259,cohen2020learning}, 
a leading system in international game competitions. Over a six-year
period, Athénan secured 57 gold medals at the Computer Olympiad~\cite{cohen2021descent,cohen2023athenan,cohen2025athenan},
tripled the record for gold medals in a single year~\cite{cohen2023athenan},
and held the title in 20 different games. Its primary competitor,
AlphaZero~\cite{silver2018general}, is based on Monte Carlo Tree
Search (MCTS)~\cite{Coulom06}. Experimental results indicate that, within
comparable computational constraints, Athénan can outperform AlphaZero
by up to a factor of 30 \cite{cohen2023minimax}.

Extending these successes to two-player stochastic games remains a
significant challenge. Introducing stochasticity fundamentally alters
the structure of the problem: the deterministic guarantees underlying
classical adversarial search no longer hold, and the evaluation of
positions must account for expected outcomes over chance events. As
a result, several key properties of the deterministic setting are
lost or weakened, including the straightforward propagation of exact
minimax values.

These characteristics make direct extensions of algorithms designed
for deterministic two-player zero-sum games insufficient. In particular,
anticipating many turns ahead becomes considerably more difficult
due to the exponential growth induced by chance nodes, often requiring
approximations that can introduce substantial biases. Moreover, algorithms
relying on precise value backpropagation must be adapted to handle
expectations, which complicates both their theoretical analysis and
practical behavior.

Furthermore, numerous algorithms have been proposed for stochastic
two-player games, yet their relative performance remains poorly understood.
It is still unclear whether a universally superior approach exists,
even in terms of average performance, and under which conditions one
method may outperform another.

In this paper, we introduce a novel search algorithm for stochastic
two-player zero-sum games with perfect information, extending the
Unbounded Minimax framework. The proposed method retains the core
principles underlying the effectiveness of Unbounded Minimax---unbounded
exploration, incremental resolution of states, and prioritization
of the most critical action sequences---while explicitly incorporating
stochastic transitions. This design helps control the combinatorial
growth induced by chance nodes and supports deeper lookahead in such
settings.

More precisely, our algorithmic contribution consists in transforming classical Expectiminimax, the standard search algorithm for stochastic games, into an anytime best-first search algorithm that generalizes Unbounded Minimax. Unlike classical Expectiminimax, which expands every chance outcome, Unbounded Expectiminimax expands only one sampled outcome per iteration, avoiding the exponential growth of the search tree due to chance and decision branching. At the same time, it preserves the exact expectiminimax backup operations rather than relying on Monte Carlo estimates as in MCTS.

We complement this contribution with an experimental evaluation on
several benchmark stochastic games, showing that the proposed approach
improves average performance compared to existing stochastic search
methods  and introduced variants.

Overall, these results contribute toward a unified perspective on
adversarial search in two-player perfect-information games, encompassing
both deterministic and stochastic settings while preserving strong
empirical performance at scale.

In summary, our contributions are:
\begin{itemize}

\item A generalization of Unbounded Minimax to stochastic games, resulting in the Unbounded Expectiminimax algorithm.
\item An enhanced version of Unbounded Expectiminimax through the adaptation of two complementary techniques to stochastic games:
\begin{itemize}
    \item the Safe Decision technique;
    \item the Completion technique.
    
\end{itemize}
\item A large-scale empirical evaluation:
\begin{itemize}
    \item across 8 stochastic games;
   \item using 30 independently trained evaluation functions per game, resulting in 30 × 30 network pairings for each domain;
    \item totaling 14,400 matches for each evaluated algorithm.
    
\end{itemize}

\end{itemize}

The remainder of this paper is organized as follows. In Section~\ref{sec:Related-Work},
we review related work, covering Unbounded Minimax and classical stochastic
game algorithms. Section~\ref{subsec:Unbounded_maxn} formalizes
our new generalized search algorithms: \textit{Unbounded Expectiminimax}
and \textit{Unbounded Expectiminimax with safe decision}. In Section~\ref{sec:Experiments},
we describe our experimental protocol, including the evaluation setup,
technical details, and the set of games considered. We then present
its results in Section~\ref{subsec:Results-of-the}. Finally, Section~\ref{sec:Conclusion}
concludes the paper and outlines directions for future work.

\section{Related Work}

\label{sec:Related-Work}
\begin{table}[t]
\begin{centering}
\begin{tabular}{|c|c|}
\hline 
{\footnotesize{}Symbols} & {\footnotesize{}Definition}\tabularnewline
\hline 
\hline 
{\footnotesize{}$\mathrm{actions}\left(s\right)$} & {\footnotesize{}action set of the state $s$ (for the current player)}\tabularnewline
\hline 
{\footnotesize{}$\mathrm{terminal\left(s\right)}$} & {\footnotesize{}true if $s$ is an end-game state}\tabularnewline
\hline 
{\footnotesize{}$a(s)$} & {\footnotesize{}new state after playing action $a$ in the state
$s$}\tabularnewline
\hline 
{\footnotesize{}$P(s,a)$} & {\footnotesize{}Game probability that action $a$ is selected in state $s$}\tabularnewline
\hline 
{\footnotesize{}$\mathrm{time}\left(\right)$} & {\footnotesize{}current time in seconds}\tabularnewline
\hline 
{\footnotesize{}$\mathrm{trunc}\left(\right)$} & {\footnotesize{}Integer truncation function}\tabularnewline
\hline 
\multirow{2}{*}{ {\footnotesize{}$S$ }} & {\footnotesize{}states of the partial game tree}\tabularnewline
 & {\footnotesize{} (and keys of the transposition table $T$)}\tabularnewline
\hline 
{\footnotesize{}$T$} & {\footnotesize{}transposition table (contains state labels as $v$)}\tabularnewline
\hline 
{\footnotesize{}$\tau$} & {\footnotesize{}search time per action}\tabularnewline
\hline 
{\footnotesize{}$t_{\max}$} & {\footnotesize{}chosen total duration of the learning process}\tabularnewline
\hline 
{\footnotesize{}$\mvs s$} & {\footnotesize{}value of state $s$ from the game search}\tabularnewline
\hline 
{\footnotesize$\mv sa$} & {\footnotesize{}value obtained after playing action $a$ in state
$s$}\tabularnewline
\hline 
{\footnotesize$\resol sa$} & {\footnotesize{}resolution value obtained after playing action $a$ in state
$s$}\tabularnewline
\hline 
{\footnotesize$\vresol sa$} & {\footnotesize{}completion value obtained after playing action $a$ in state
$s$}\tabularnewline
\hline 
{\footnotesize$\select sa$} & {\footnotesize number of times the action $a$ is selected in state
$s$}\tabularnewline
\hline 
\multirow{2}{*}{{\footnotesize{}$\hadapt(s)$}} & {\footnotesize{}adaptive evaluation function (of}\tabularnewline
 & {\footnotesize{} non-terminal states ; first player point of view)}\tabularnewline
\hline 
\multirow{2}{*}{{\footnotesize{}$\hterminal(s)$}} & {\footnotesize{}evaluation of terminal states, e.g.}\tabularnewline
 & {\footnotesize{} gain game (first player point of view)}\tabularnewline
\hline 
\multirow{5}{*}{ 

\parbox{1.15cm}{ \center
{\footnotesize{}search \\ ($s$, $S$, $T$) }}
}

 & {\footnotesize{}a search algorithm (it extends the game tree}\tabularnewline
 & {\footnotesize{} from $s$, by adding new states in $S$ and}\tabularnewline
 & {\footnotesize{}labeling its states by a value $\mvs s$, stored
in}\tabularnewline
 & {\footnotesize{}$T$, using $\hadapt$ as evaluation of non-terminal}\tabularnewline
 & {\footnotesize{}leaves and $\hterminal$ as evaluation of terminal
states}\tabularnewline
\hline 
\end{tabular}{\footnotesize}{\footnotesize\par}
\par\end{centering}
\caption{Index of symbols}\label{tab:Index-of-symbols}
\end{table}

\subsection{Best Action Searching in Game Trees}

The vast majority of game-search algorithms rely on representing a
game as a tree, where nodes correspond to game states and edges correspond
to actions that allow transitions from one state to another. The terminal
nodes represent end-game states and are assigned a value according
to the final game outcome. Typically $v\in\left\{ -1,0,1\right\} $
where $v=1$ means the first player wins, $v=0$ means draw, and $v=-1$
means the first player loses (and thus the second player wins). More
generally, $v\in\mathbb{R}$ and $v$ is the score of the first player.
This valuation can be propagated to the other nodes under the assumption
that both players play optimally. However, this theoretical valuation
is generally intractable in practice, as game trees represent enormous
state spaces. Consequently, algorithms must estimate these values
or the corresponding strategies to make practical game-playing decisions.

\subsection{Base Stochastic Search: Expectiminimax}

\label{subsec:Classical-adversarial-search}

The Expectiminimax algorithm~\cite{michie1966game} is the canonical
approach for stochastic zero-sum two-player games: it constructs a
game tree to a fixed depth, evaluates the leaf nodes with a heuristic
evaluation function, and propagates these values back up the tree
under the assumption that the first player maximizes its payoff, that
the second player minimizes the first player payoff, and that stochastic
events follow the probabilities defined by the game. Expectiminimax
remains the simplest and most fundamental framework for reasoning
about stochastic two-player adversarial behavior.

When upper and lower bounds on the values of horizon states are known,
a pruning generalizing the alpha-beta technique can be used; this
technique is called {*}-minimax~\cite{ballard1983minimax}. Our study
is not situated within this specific setting.

\subsection{Monte-Carlo Tree Search}

\label{subsec:Monte-Carlo-Tree-Search}

Monte Carlo Tree Search (MCTS) \cite{browne2012survey,Coulom06} reframed
game search for domains where hand-crafted evaluation functions are
inadequate. Instead of evaluating every leaf node with a heuristic,
MCTS performs randomized simulations (playouts) and uses their outcomes
to guide a best-first expansion of the search tree through four iterative
steps: selection, expansion, simulation, and backpropagation. MCTS
has achieved remarkable success across a wide range of board and real-time
games. $\mcts{}$ is parameterized by a constant $C$ which needs
to be tuned (theoretical value $\sqrt{2}$).

It has also been modified in various ways to incorporate knowledge
to improve its performance by biasing its selections and/or simulations~\cite{browne2012survey}.
The simplest method is to replace state evaluation by playouts with
heuristic evaluation~\cite{ramanujan2011trade}. We denote this variant
by $\mctsh{}$.

Using MCTS in stochastic games introduces chance nodes. Two generalisations
are commonly considered. The first proceeds with selection until a
state is reached that is not present in the tree, regardless of whether
it is a chance or a decision node. The second continues exploration
until a previously unseen decision node is encountered~\cite{openspiel:mcts}. In this variant,
chance nodes are always fully expanded according to the game’s probability
distribution and are not evaluated by the evaluation function (when
such a function is used). We denote this variant as $\mctshd{}$.

Open-loop MCTS~\cite{perez2015open} is a variant of Monte Carlo
Tree Search that evaluates action sequences rather than state-action
pairs, thereby avoiding explicit dependence on exact state identities.
It is particularly suited to stochastic or partially observable environments,
where different states can lead to similar action distributions, and
it improves scalability by aggregating statistics over action histories
instead of full state representations. We denote this variant using
heuristic evaluation by $\mctsh{\text{open}}$.

\subsection{Unbounded Minimax and Safe Decision }

\label{subsec:Athenan}

\subsubsection{Unbounded Minimax}

Unbounded (Best-First) Minimax \cite{cohen2025study,korf1996best,cohen2025little,JMLR:v27:25-2259,cohen2021completeness}
is an extension of the classical Minimax, the base algorithm for two-player
zero-sum games with perfect information. In its standard form, Minimax
explores the game tree uniformly up to a fixed depth $d$, assuming
both players act optimally within this search horizon, one maximizing
the payoff and the other minimizing it. This approach, however, quickly
becomes intractable as $d$ increases, due to the exponential growth
of the search space.

To overcome this limitation, Unbounded Minimax replaces depth-limited
exploration with a best-first, non-uniform search strategy. Instead
of expanding the entire frontier, it focuses selectively on the most
promising parts of the search space, enabling significantly deeper
local exploration. Concretely, the algorithm identifies the currently
best sequence of actions, extends it by one step, and iteratively
repeats this process. Unlike standard Minimax, it does not rely on
a predefined depth parameter: the search progresses incrementally
and can be stopped at any time while still providing a valid solution.

\subsubsection{Safe Unbounded Minimax}

Unbounded Minimax was later extended with the safe decision technique~\cite{cohen2025study,JMLR:v27:25-2259},
which modifies the final action-selection rule while leaving the underlying
search procedure unchanged. In the standard version, the selected
action is the one leading to the child state with the highest evaluated
value. In contrast, the safe decision strategy selects the action
deemed the safest, defined as the one most frequently chosen during
the search.

More precisely, the selected action maximizes the number of times
    that action has been preferred at the root throughout the search process. Given
that the exploration follows a best-first strategy, this action corresponds
to the one most often identified as optimal during successive iterations.
This mechanism improves robustness to evaluation inaccuracies. Additionally,
it favors the action for which the algorithm has explored the deepest
and most extensively developed subtree, thereby providing greater
insight into its potential future outcomes.

\subsubsection{Unbounded Minimax with Completion}

Unbounded Minimax has also been enhanced through the completion technique~\cite{cohen2021completeness,JMLR:v27:25-2259}.
This extension introduces two key improvements. First, it prunes actions
leading to solved states, i.e., states whose exact value is known
and cannot be altered by further exploration, thereby preventing the
search from being trapped in suboptimal fixed points. Second, it refines
the final decision rule by systematically prioritizing actions that
lead to solved winning states and, whenever possible, avoiding those
that lead to solved losing states.

\section{Contributions}

\subsection{A New Search Algorithm for Stochastic Games}\label{subsec:Unbounded_maxn}

In this section, we introduce our generalization of Unbounded Minimax
to the stochastic setting. We call this algorithm \textit{Unbounded
}\emph{Expectiminimax.} Like Unbounded Minimax for deterministic games,
this algorithm iteratively extends the a priori best action sequence
as long as search time remains, except that in added chance states,
the next state is chosen at random and that the values of states are
no longer the minimax value but the expectiminimax value (of the partial
game tree). 

In contrast to classic Expectiminimax, our algorithm mitigates the
combinatorial explosion induced by chance nodes. It expands only a
single sampled successor at each chance node per iteration, progressively
refining value estimates through repeated sampling. This approach
combines the scalability of sampling-based methods, which avoid exhaustive
expansion, with the principled backup mechanism of Expectiminimax,
which preserves correctness when exact evaluations are available.

At each iteration, the algorithm performs a recursive selection from
the root (the current game state), descending the tree according to
the following decision rules: in max nodes (first-player states),
it selects the child with maximal value; in min nodes (second-player
states), the child with minimal value; and in chance nodes, a successor
is sampled according to the probability distribution defined by the
game. This process continues until a leaf of the current partial tree
is reached, at which point expansion occurs: the newly encountered
state is added to the search tree. Then, value updates are propagated
backward.

During backpropagation, node values are updated according to their
type: max nodes take the maximum over their children, min nodes the
minimum, and chance nodes the expectation over their children weighted
by their transition probabilities. This procedure is repeated from
the root for the available search budget, yielding an anytime algorithm
that progressively improves its decision quality over time.

The completion technique is generalized as follows: at the end of
the search, always play an action that is resolved as winning with
probability $1$, and, if possible, never play an action that is resolved
as losing with probability $1$. Thus, an action resolved as winning
with probability $p<1$ is not necessarily played. A state is considered
resolved if all its children are resolved, or if it has a winning
action for the current player with probability $1$. Finally, during
the search, a resolved action is never explored. An action $a$ in
a state $s$ is marked as resolved when $\resol sa=1$ (otherwise
$\resol sa=0$). An action $a$ in a state $s$ is marked as resolved
winning with probability $1$ when $\vresol sa=1$. An action $a$
in a state $s$ is marked as resolved losing with probability $1$
when $\vresol sa=-1$.

The Unbounded Expectiminimax algorithm for stochastic games (with
completion and optionally with safe decision) is described in Algorithm~\ref{alg:umaxn}.

\begin{algorithm}
\centering{}\includegraphics[scale=0.57]{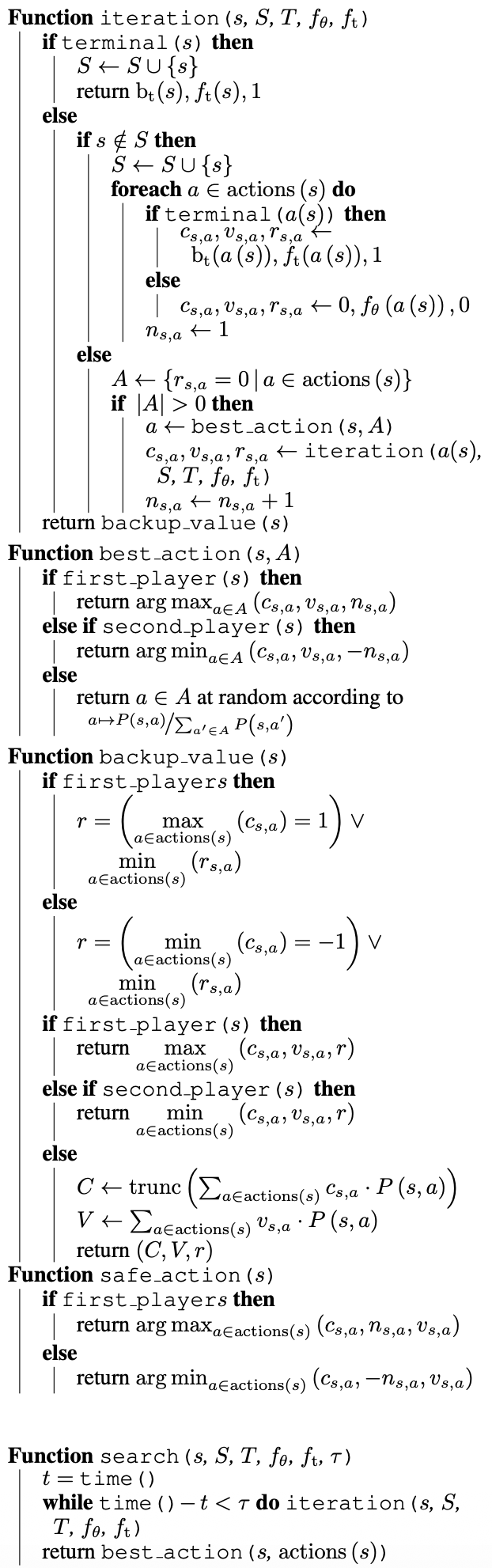}\caption{\textbf{Unbounded }Expectiminimax (with completion). This algorithm
	computes the best action to play in the generated non-uniform partial
	game tree starting from the root state $s$ with $\protect\tmax$
	as search time. See Table~\ref{tab:Index-of-symbols} for the definitions
	of symbols. Note: tuples are lexicographically ordered. To obtain
	the safe decision version: replace best\_action() in the last line
	of search() by safe\_action().}\label{alg:umaxn}
\end{algorithm}

\subsection{Experimental Study of Stochastic Searches}\label{sec:Experiments}

We now present the experiments conducted to evaluate the contributions
of this paper. We will present its results in the following section
(Section~\ref{subsec:Results-of-the}). 

We begin by presenting the experimental protocol (Section~\ref{subsec:Evaluation-Protocol}).
Then we explain how the evaluation functions used by the search algorithms
were generated (Section~\ref{subsec:apprentissage-fonctions-evaluation}).
Next, we describe the computer used (Section~\ref{subsec:computer_used}),
and finally present the other details of the experiment (Section~\ref{subsec:Other-Details-of}).

\subsubsection{Evaluation Protocol}\label{subsec:Evaluation-Protocol}

We experimentally compare the following algorithms: (classic) Expectiminimax,
$\mcts{}$, $\mctsh{}$, $\mctsh{\text{open}}$, $\mctshd{}$, Unbounded
Expectiminimax with classic decision and with safe decision.

We use standard Expectiminimax as benchmark adversary using a lossless
parallelization procedure, called Child Batching \cite{cohen2025study,JMLR:v27:25-2259}
(child Batching consists in evaluating the children states of states
in parallel by the neural network). We evaluate each algorithm on
the following 8 games: Azul, Kingdomino, Mighty Duel Kingdomino, Can't
Stop, Backgammon, EinStein würfelt nicht!, Shogun, and Martian Dice.
Corresponding descriptions are in Appendix.

Each algorithm faces the benchmark adversary $2$ times: once in first
player and once in second player. These matches are repeated $30\times30$
times: we use a set of 30 evaluation functions per game in order to
assess the algorithms statistically. More precisely, each algorithm
is evaluated for each evaluation function $f_{i}$ ($i\in\left\{ 0,\ldots,29\right\} $)
that it uses and for each opponent evaluation function $f_{j}$ ($j\in\left\{ 0,\ldots,29\right\} $).
In other words, for all $i,j\in\left\{ 0,\ldots,,29\right\} $, the
evaluated algorithm using $f_{i}$ confronts in a same match the following
opponent: the benchmark adversary with the evaluation $f_{j}$. This
corresponds to 1800 matches per game for each evaluated algorithm. The performance
evaluation of an algorithm for a specific game is therefore based
on this number of matches. The performance of a search algorithm is
the average of the binary scores obtained at the end of its matches
(1 for a victory, 0 for a draw and -1 for a defeat). 

Each search algorithm uses the same search time: 10 seconds per action
(order of magnitude of search time in competition). 

Each algorithm is evaluated with and without Completion. For MCTS
and classic Expectiminimax, this means always playing a solved action
with a 100\% winning probability and never a solved action with a
100\% losing probability after the search. Unbounded Expectiminimax,
in addition, never selects a solved action during the search. This
sub-technique is irrelevant for classic Expectiminimax and significantly
reduces performance for Monte Carlo Tree Search \cite{winands2008monte}
(we have also verified this in the context of this experiment : it
decreases by $25\%$ the performance).

The evaluated algorithms do not use Child Batching, unlike their
opponent. We do not use it because Child Batching cannot be applied
to MCTS algorithms, as they do not evaluate all children of a state
sequentially. Although some form of parallelization is possible, it
comes at the cost of performance, and parallelization is not the focus
of this paper. To ensure a fair comparison, we therefore disable Child
Batching for all evaluated algorithms.

Note, however, that Expectiminimax-based algorithms, including both
the classical algorithm and our unbounded variants, can make use of
Child Batching. This improves their absolute performance, but does
not alter their relative performance. Child Batching increases the
performance of Expectiminimax algorithms by around $6\%$. 

\subsubsection{Evaluation Functions Training}\label{subsec:apprentissage-fonctions-evaluation}

We now present how the evaluation functions used in the experiments
of this paper were learned.

The evaluation functions were generated using Athénan \cite{JMLR:v27:25-2259}
; the reinforcement learning algorithm, by replacing its core component
Descent Minimax, with its generalization for stochastic games: Descent
Expectiminimax~\cite{cohen2023learning}. 

Athénan involves sequentially performing self-play matches using the
learned evaluation function to guide the search process, deciding
which actions to take during the matches and how to explore the search
tree. The data encountered (the associations states-values) is then
learned at the end of each match to update the evaluation function.
This thus forms a self-improving loop.

The Athénan hyperparameters used to generate each evaluation function
are the following: search time per action $t=2s$, batch size $B=3000$,
duplication factor $\delta=3$, memory size $M=50$, the selection
action method is still the ordinal law but its parameter is $\sqrt{X}$
with $X$ a uniform random variable in {[}0,1{]}, neural network optimizer:
Adam~\cite{kingma2014adam} with $\lambda=0.0001$ as learning rate.
The neural network architecture used is as follow: a convolutional
layer, $8$ residual blocks~\cite{he2016deep} (containing 2 convolutional
layers), a flatten layer, two dense layers with $N$ neurons 
(see Table~\ref{tab:F_N_values_ExIt_Ath=0000E9nan})
, and a final dense layer with
$1$ neuron (the output). Each convolutional layer has $C$ channels
(see Table~\ref{tab:F_N_values_ExIt_Ath=0000E9nan})
 The activation function
is the ReLU.

Each of the evaluation functions is the result of 10 days of training.

\begin{table}[t]
\centering{}%
\begin{tabular}{|c|c|c|}
\hline 
 & $C$ & $N$\tabularnewline
\hline 
\hline 
\emph{EinStein würfelt nicht!} & $131$ & $1097$\tabularnewline
\hline 
Backgammon & $131$ & $245$\tabularnewline
\hline 
Kingdomino & $125$ & $1103$\tabularnewline
\hline 
Mighty Duel Kingdomino & $129$ & $645$\tabularnewline
\hline 
Azul & $128$ & $1106$\tabularnewline
\hline 
Martian Dice & $228$ & $1033$\tabularnewline
\hline 
Shogun & $132$ & $477$\tabularnewline
\hline 
Can't stop & $131$ & $156$\tabularnewline
\hline 
\end{tabular}\caption{Number of convolutional channels $C$ and hidden dense neurons $N$
used by Athénan for each studied game.}\label{tab:F_N_values_ExIt_Ath=0000E9nan}
\end{table}

\subsubsection{Computer Used}\label{subsec:computer_used}

We provide details of the machines used in the experiments described
in this article.

We used the LUMI-G partition of the LUMI Supercomputer. A process
uses 7 AMD Trento CPU with 64 GB of RAM and 1 AMD MI250 GPU with 64
GB of RAM.

\subsubsection{Other Details of the Experiment}\label{subsec:Other-Details-of}

We give for each algorithm: the average performance over all games,
the associated 95\% stratified bootstrapping confidence interval (lower
and upper bound) and the performance for each game with its 95\% confidence
radius. The percentages detailing performance for each game have been
rounded to the nearest percent to reduce size and improve readability
of data tables.

Algorithms based on a fixed search depth (classic Expectiminimax)
use iterative deepening~\cite{korf1985depth} to be parameterized
by a search time. 

All algorithms use transposition tables~\cite{greenblatt1988greenblatt}.
This allows the game tree to be transformed into a directed acyclic
graph, which therefore simplifies the state space.

The evaluation functions for $\mctsh{}$ need to take values in $[0,1]$.
We use the following normalization to obtain such evaluation functions,
transforming an evaluation function $f$ with values in $\mathbb{R}$
into an evaluation function $f_{\mathrm{n}}$ with values in $[0,1]$:

\[
f_{\mathrm{n}}(s)=\frac{\max\left(\min\left(f\left(s\right),M\right),m\right)-m}{M-m}
\]
with $s$ a game state, $f$ the state evaluation function, $M$ the
maximum practical value for $f$, $m$ the minimum practical value
of $f$. More formally, $M=\max V$ and $m=\min V$ with $V=\liste{f\left(s'\right)}{s'\in S}$
where $S$ is a set of states from matches played using $f$ generated
during 24 hours.

\subsection{Results of the Experimental Study}\label{subsec:Results-of-the}

We now present the results of the experiment. In Section~\ref{subsec:First-stage-of},
we present the performance of algorithms with a medium search time
per action. In Section~\ref{subsec:Short-time-experiment},
we present the performance of algorithms with a short search time
per action. 

\subsubsection{Comparison with a Medium Search Time}\label{subsec:First-stage-of}

We thus evaluate the following algorithms: $\mcts{}$, $\mctsh{}$,
$\mctsh{\text{open}}$, $\mctshd{}$ and Unbounded Expectiminimax
with and without Safe Decision, and classic
Expectiminimax. The algorithms performances are described in Table~\ref{tab:res_sans-batch-mean}
(mean over the games) and Table~\ref{tab:res_sans-batch-details}
(details).

Overall, Unbounded Expectiminimax with and without safe decision achieves
similar results, indistinguishable within the confidence interval
(both with and without completion). This contrasts with the deterministic
case, where safe decision substantially improves performance. In contrast,
as in the deterministic setting, Completion significantly improves
performance, by approximately $8\%$. The Unbounded Expectiminimax
algorithms generally outperform all other algorithms. Classic Expectiminimax
achieves results comparable to $\mctsh{}$ and $\mctsh{'}$
without Completion but it is stronger with completion. Evaluating
chance nodes with $\mctsh{}$ has only a very limited impact on performance.
On the other hand, the open-loop variant performs significantly worse.
The worst-performing algorithm overall is $\mcts{}$, i.e. without
the use of learned knowledge (showing that learning is indeed beneficial
and has a strong impact on performance).

We now examine the results in greater detail on a game-by-game basis,
highlighting the points where the observed behavior diverges from
the overall trend.

In Azul without using Completion, Monte Carlo approaches using learned
knowledge perform best. In particular, $\mctsh{}$ achieves the best
results. However, adding Completion greatly improves Unbounded Expectiminimax,
which is not the case with MCTS, and Unbounded Expectiminimax with Completion
achieves the performance of $\mctsh{}$.

In Backgammon, Martian Dice, and Einstein würfelt nicht!, completion
has no measurable impact.

In Azul, safe decision significantly improves performance in this
setting, whereas it significantly decreases performance in Einstein
würfelt nicht!.

\begin{table*}[t]
\begin{centering}
{\tiny{}%
\begin{tabular}{|c|c|c|c|c|c|c|c|c|}
\hline 
 & {\tiny Azul   } & {\tiny Backgammon } & {\tiny Can't Stop } & {\tiny Einstein würfelt nicht!} & {\tiny Kingdomino } & {\tiny Mighty Duel Kingdomino} & {\tiny Martian Dice } & {\tiny Shogun }\tabularnewline
\hline 
\hline 
{\tiny Unbounded Expectiminimax} & {\tiny 4 \textpm 4 } & {\tiny{} \textbf{2} \textpm 4 } & {\tiny{} \textbf{6} \textpm 3 } & {\tiny{} -3 \textpm 4 } & {\tiny{} \textbf{19} \textpm 4 } & {\tiny{} \textbf{9} \textpm 4 } & {\tiny{} \textbf{-1} \textpm 4 } & {\tiny{} \textbf{19} \textpm 4 }\tabularnewline
\hline 
{\tiny Safe Unbounded Expectiminimax} & {\tiny \textbf{20} \textpm 4 } & {\tiny{} -4 \textpm 4 } & {\tiny{} 1 \textpm 3 } & {\tiny{} -23 \textpm 4 } & {\tiny{} \textbf{25} \textpm 4 } & {\tiny{} \textbf{10} \textpm 4 } & {\tiny{} -7 \textpm 4 } & {\tiny{} \textbf{18} \textpm 4 }\tabularnewline
\hline 
{\tiny classic Expectiminimax with Completion} & {\tiny -7 \textpm 4} & {\tiny{} -4 \textpm 4} & {\tiny{} \textbf{4} \textpm 3 } & {\tiny{} \textbf{0} \textpm 4 } & {\tiny{} -3 \textpm 4} & {\tiny{} 2 \textpm 4 } & {\tiny{} \textbf{0} \textpm 4 } & {\tiny{} 0 \textpm 4}\tabularnewline
\hline 
{\tiny Safe Unbounded Exp. without Completion} & {\tiny -8 \textpm 4 } & {\tiny{} -3 \textpm 4 } & {\tiny{} -5 \textpm 3 } & {\tiny{} -3 \textpm 4 } & {\tiny{} 7 \textpm 4 } & {\tiny{} -2 \textpm 4 } & {\tiny{} \textbf{-2} \textpm 4 } & {\tiny{} 7 \textpm 4 }\tabularnewline
\hline 
{\tiny Unbounded Exp. without Completion} & {\tiny -18 \textpm 4 } & {\tiny{} \textbf{0} \textpm 4 } & {\tiny{} -2 \textpm 3 } & {\tiny{} \textbf{-2} \textpm 4 } & {\tiny{} 1 \textpm 4 } & {\tiny{} -2 \textpm 4 } & {\tiny{} -6 \textpm 4 } & {\tiny{} 7 \textpm 4 }\tabularnewline
\hline 
{\tiny classic Expectiminimax} & {\tiny -13 \textpm 4 } & {\tiny{} -7 \textpm 4 } & {\tiny{} -4 \textpm 3 } & {\tiny{} \textbf{2} \textpm 4 } & {\tiny{} 3 \textpm 4 } & {\tiny{} -6 \textpm 4 } & {\tiny{} \textbf{1} \textpm 4 } & {\tiny{} -26 \textpm 4}\tabularnewline
\hline 
{\tiny$\mctsh{}$ with Completion} & {\tiny \textbf{17} \textpm 4} & {\tiny{} -13 \textpm 4 } & {\tiny{} 2 \textpm 3 } & {\tiny{} -7 \textpm 4} & {\tiny{} 1 \textpm 4 } & {\tiny{} -1 \textpm 4} & {\tiny{} -7 \textpm 4} & {\tiny{} -48 \textpm 3}\tabularnewline
\hline 
{\tiny$\mctsh{}$} & {\tiny \textbf{20} \textpm 4} & {\tiny{} -10 \textpm 4 } & {\tiny{} 1 \textpm 3 } & {\tiny{} \textbf{2} \textpm 4 } & {\tiny{} -6 \textpm 4} & {\tiny{} -1 \textpm 4} & {\tiny{} \textbf{-2} \textpm 4} & {\tiny{} -62 \textpm 3}\tabularnewline
\hline 
{\tiny$\mctsh{'}$} & {\tiny 16 \textpm 4} & {\tiny{} -15 \textpm 4 } & {\tiny{} \textbf{7} \textpm 3 } & {\tiny{} \textbf{1} \textpm 4 } & {\tiny{} -7 \textpm 4} & {\tiny{} 2 \textpm 4 } & {\tiny{} \textbf{-1} \textpm 4} & {\tiny{} -72 \textpm 3}\tabularnewline
\hline 
{\tiny$\mctsh{\text{open}}$} & {\tiny 15 \textpm 4} & {\tiny{} -26 \textpm 4 } & {\tiny{} \textbf{4} \textpm 3 } & {\tiny{} -4 \textpm 4} & {\tiny{} -23 \textpm 4 } & {\tiny{} -17 \textpm 4 } & {\tiny{} -21 \textpm 4 } & {\tiny{} -82 \textpm 2}\tabularnewline
\hline 
{\tiny$\mcts{}$} & {\tiny -97 \textpm{} 1 } & {\tiny{} -94 \textpm{} 1 } & {\tiny{} -7 \textpm{} 4} & {\tiny{} -10 \textpm{} 4 } & {\tiny{} -23 \textpm{} 4} & {\tiny{} -78 \textpm{} 3} & {\tiny{} -40 \textpm{} 4 } & {\tiny{} -97 \textpm{} 1 }\tabularnewline
\hline 
\end{tabular}}{\tiny}{\tiny\par}
\par\end{centering}
\caption{Binary scores for all games of the evaluated algorithms without batching
against Expectiminimax with batching (All algorithms have 10s per action).}\label{tab:res_sans-batch-details}
\end{table*}

\begin{table}
\begin{centering}
{\small{}%
\begin{tabular}{|c|c|c|c|}
\hline 
 & {\small mean} & {\small lower bound} & {\small upper bound}\tabularnewline
\hline 
\hline 
{\small Unbounded E +C} & {\small \textbf{6.77} } & {\small 5.61} & {\small 7.93}\tabularnewline
\hline 
{\small Safe Unbounded E +C} & {\small \textbf{5.01}} & {\small 3.87} & {\small 6.17}\tabularnewline
\hline 
{\small Expectiminimax +C} & {\small -0.99} & {\small -2.15} & {\small 0.18}\tabularnewline
\hline 
Safe {\small Unbounded E} & {\small -1.43} & {\small -2.59} & {\small -0.26}\tabularnewline
\hline 
{\small Unbounded E } & {\small -2.7} & {\small -3.87} & {\small -1.55}\tabularnewline
\hline 
{\small Expecti.} & {\small -6} & {\small -7.18} & {\small -4.82}\tabularnewline
\hline 
{\small$\mctsh{}$ +C} & {\small -6.82} & {\small -7.95} & {\small -5.68}\tabularnewline
\hline 
{\small$\mctsh{}$} & {\small -7.24} & {\small -8.37} & {\small -6.11}\tabularnewline
\hline 
{\small$\mctsh{'}$} & {\small -8.49} & {\small -9.62} & {\small -7.37}\tabularnewline
\hline 
{\small$\mctsh{\text{open}}$} & {\small -19.3} & {\small -20.39} & {\small -18.18}\tabularnewline
\hline 
{\small$\mcts{}$} & {\small -55.73} & {\small -56.61} & {\small -54.84}\tabularnewline
\hline 
\end{tabular}}{\small}{\small\par}
\par\end{centering}
\caption{Average binary scores over all games of the evaluated algorithms without
batching against classic Expectiminimax with batching (E: Expectiminimax ; +C: with Completion).  All algorithms have 10s per action.}\label{tab:res_sans-batch-mean}
\end{table}

\subsubsection{Comparison with a Short Search} \label{subsec:Short-time-experiment}

In this section, we repeat the main experiment of the paper using
a short search time (1 second).

We thus evaluate the following algorithms in this modified context:
$\mcts{}$, $\mctsh{}$, $\mctsh{\text{open}}$, $\mctshd{}$ and
Unbounded Expectiminimax with and without Safe Decision,
and classic Expectiminimax. 
The algorithms’ performances are summarized in Table~\ref{tab:res_sans-batch-mean-1}, which reports the average performance across games. Detailed results for each individual game are provided in Appendix Table~\ref{tab:res_sans-batch-details-1}.

We recover the main conclusions observed in the 10-second setting
(Safe Unbounded Expectiminimax without Completion is more efficient
than classic Expectiminimax, $\mctsh{}$ and $\mctsh{'}$
which are more efficient than $\mctsh{\text{open}}$, itself more
efficient than $\mcts{}$; Completion improves performance).

There are, however, two notable global differences:
\begin{itemize}
\item Safe decision improves performance, as in the deterministic case (although
the effect is less pronounced).
\item $\mctsh{}$ outperforms $\mctsh{'}$: evaluating chance
states therefore improves performance.
\end{itemize}
Regarding game-specific observations that diverge from the overall
behavior, we observe several results that are consistent with those
reported previously in the 10-second setting:
\begin{itemize}
\item In Azul, safe decision significantly improves performance, whereas
it significantly decreases performance in Einstein würfelt nicht!.
\item In Backgammon, Martian Dice, and Einstein würfelt nicht!, completion
still has no measurable impact.
\end{itemize}
There are, however, some game-specific differences:
\begin{itemize}
\item In Can’t Stop, without Completion, $\mctsh{}$ is the best-performing
algorithm.
\item In the 10-second setting for Azul, without Completion, $\mctsh{}$
was the best-performing algorithm. This is no longer the case here,
as Unbounded Expectiminimax with safe decision achieves the best results.
\item In Kingdomino, Mighty Duel Kingdomino, and Shogun, safe decision significantly
improves performance.
\end{itemize}

\begin{table}[t]
\begin{centering}
{\small{}%
\begin{tabular}{|c|c|c|c|}
\hline 
 & {\small mean} & {\small lower bound} & {\small upper bound}\tabularnewline
\hline 
\hline 
{\small Unbounded E +C} & {\small 1.51 } & {\small 0.35} & {\small 2.66}\tabularnewline
\hline 
{\small Safe Unbounded E +C} & {\small \textbf{4.69}} & {\small 3.55} & {\small 5.84}\tabularnewline
\hline 
{\small Safe Unbounded E} & {\small 1.72} & {\small 0.57} & {\small 2.87}\tabularnewline
\hline 
{\small Unbounded E} & {\small -3.81} & {\small -4.98} & {\small -2.65}\tabularnewline
\hline 
{\small  Expectiminimax +C} & {\small -3.84} & {\small -5.01} & {\small -2.7}\tabularnewline
\hline 
{\small Expectiminimax} & {\small -5.12} & {\small -6.27} & {\small -3.97}\tabularnewline
\hline 
{\small$\mctsh{}$ +C} & {\small -9.97} & {\small -11.12} & {\small -8.82}\tabularnewline
\hline 
{\small$\mctsh{}$} & {\small -11.09} & {\small -12.24} & {\small -9.96}\tabularnewline
\hline 
{\small$\mctsh{'}$} & {\small -16.53} & {\small -17.66} & {\small -15.4}\tabularnewline
\hline 
{\small$\mctsh{\text{open}}$} & {\small -24.47} & {\small -25.55} & {\small -23.37}\tabularnewline
\hline 
{\small$\mcts{}$} & {\small -70.22} & {\small -70.96} & {\small -69.49}\tabularnewline
\hline 
\end{tabular}}{\small\par}
\par\end{centering}
\begin{centering}
{\small}{\small\par}
\par\end{centering}
\caption{Average binary scores over all games of the evaluated algorithms without
batching against classic Expectiminimax with batching (E: Expectiminimax ; +C: with Completion). All algorithms have 1s per action.}\label{tab:res_sans-batch-mean-1}
\end{table}

\section{Discussion}\label{sec:Discussion}

Beyond the empirical results, the design of Unbounded Expectiminimax highlights several key differences with both classical Expectiminimax and Monte-Carlo Tree Search. These differences provide insights into the conditions under which Unbounded Expectiminimax may be particularly effective and suggest possible explanations for its observed performance.

Rather than approximating expectiminimax values through stochastic rollouts, Unbounded Expectiminimax preserves the exact expectiminimax backup operation. As a result, the quality of the search depends primarily on the accuracy of the evaluation function rather than on Monte-Carlo estimation variance. In the ideal case of a perfect evaluation function, the optimal action can be identified after a single iteration, whereas MCTS must still reduce rollout variance through repeated simulations.

Another important consequence is that Unbounded Expectiminimax naturally inherits the best-first search behavior of Unbounded Minimax. Instead of distributing the search effort uniformly across the game tree, it concentrates computation on the most promising branches. This selective expansion enables the algorithm to reach strategically deeper positions under a fixed computational budget, provided that the evaluation function is sufficiently informative.

The proposed tree construction strategy also changes the computational characteristics of expectiminimax search. Classical Expectiminimax expands all successors of every chance node, leading to exponential growth of the search tree and requiring iterative deepening to provide intermediate decisions. In contrast, Unbounded Expectiminimax expands only one sampled outcome at each iteration, 
making the number of newly expanded nodes per iteration independent of the branching factors of decision and chance nodes, while remaining an anytime algorithm without additional computation.

Finally, unlike MCTS, Unbounded Expectiminimax does not rely on an explicit exploration–exploitation trade-off or on Monte-Carlo value estimation. The search is entirely driven by backed-up expectiminimax values, eliminating the need for exploration constants or other search hyperparameters. This makes the search procedure parameter-free and simplifies its practical application across different stochastic games without domain-specific tuning.

Overall, these properties suggest that Unbounded Expectiminimax is particularly well suited to stochastic games where reasonably accurate evaluation functions are available. In such settings, replacing exhaustive chance-node expansion with repeated sampling preserves the strengths of Expectiminimax while substantially improving its scalability and anytime behavior.

Our experimental results are consistent with this analysis and show that evaluation functions sufficiently informative to make Unbounded Expectiminimax the best-performing algorithm can be obtained with only a few days of computation on a GPU.

\section{Conclusion}\label{sec:Conclusion}


In this paper, we investigated search algorithms for stochastic perfect-information games. We generalized Unbounded Minimax with Safe Decision and Completion, a state-of-the-art approach for deterministic two-player zero-sum perfect-information games, to the stochastic setting, resulting in Unbounded Expectiminimax and its variants.

We evaluated these algorithms through a large-scale empirical study against the main stochastic search algorithms from the literature. Our results show that our generalizations achieve the best average performance for medium search times. Unlike in deterministic settings, however, Safe Decision does not consistently provide a significant improvement over the classical decision rule. Their performances are generally close, and in one game the classical decision rule performs substantially better. Completion, in contrast, consistently improves performance.

We also evaluated the algorithms under a shorter search budget (1 second). The results are consistent with those obtained for medium search times (10 seconds), with the difference that Safe Decision provides additional improvements in this setting.

Across both search budgets, Unbounded Expectiminimax achieves the best average performance among all evaluated stochastic search algorithms. This work extends the unbounded best-first search paradigm from deterministic to stochastic perfect-information games, providing a unified framework covering both settings.

Overall, our results suggest that Unbounded Expectiminimax with Safe Decision is a strong default choice for stochastic perfect-information games when an accurate evaluation function is available or can be learned, for example through several days of GPU-based training. In some specific cases, such as particular games or search budgets, the classical decision rule can still provide competitive or superior performance.

Future work will investigate the extension of this framework to multiplayer and imperfect-information games.


\bibliography{jeux}

\clearpage{}

\section*{Appendix}

We present in this document the details of the experiments.



Table~\ref{tab:res_sans-batch-details-1} detailed the results of the Section  \ref{subsec:Short-time-experiment} for each individual game.

In Section~\ref{subsec:tuning}, we present the tuning of parameterized
search algorithms.

In Section~\ref{subsec:Games}, we present the games used as benchmarks
for the experiments in this article.

\begin{table*}[t]
\begin{centering}
{\tiny{}%
\begin{tabular}{|c|c|c|c|c|c|c|c|c|}
\hline 
 & {\tiny Azul} & {\tiny Backgammon} & {\tiny Can't Stop} & {\tiny Einstein würfelt nicht!} & {\tiny Kingdomino} & {\tiny Mighty Duel Kingdomino} & {\tiny Martian Dice} & {\tiny Shogun}\tabularnewline
\hline 
\hline 
{\tiny Unbounded Expectiminimax} & {\tiny -6 \textpm 4 } & {\tiny{} -1 \textpm 4 } & {\tiny{} \textbf{2} \textpm 3 } & {\tiny{} \textbf{3} \textpm 4 } & {\tiny{} 1 \textpm 4 } & {\tiny{} -10 \textpm 4 } & {\tiny{} \textbf{0} \textpm 4 } & {\tiny{} \textbf{22} \textpm 4 }\tabularnewline
\hline 
{\tiny Safe Unbounded Expectiminimax} & {\tiny \textbf{14} \textpm 4 } & {\tiny{} \textbf{3} \textpm 4 } & {\tiny{} \textbf{3} \textpm 3 } & {\tiny{} -10 \textpm 4 } & {\tiny{} \textbf{10} \textpm 4 } & {\tiny{} \textbf{7} \textpm 4 } & {\tiny{} \textbf{-2} \textpm 4 } & {\tiny{} 13 \textpm 4 }\tabularnewline
\hline 
{\tiny Safe Unbounded Exp. without Completion} & {\tiny -1 \textpm 4 } & {\tiny{} \textbf{4} \textpm 4 } & {\tiny{} -5 \textpm 3 } & {\tiny{} -2 \textpm 4 } & {\tiny{} \textbf{8} \textpm 4 } & {\tiny{} \textbf{3} \textpm 4 } & {\tiny{} -4 \textpm 4 } & {\tiny{} 9 \textpm 4 }\tabularnewline
\hline 
{\tiny Unbounded Exp. without Completion} & {\tiny -20 \textpm 4 } & {\tiny{} \textbf{1} \textpm 4 } & {\tiny{} -2 \textpm 3 } & {\tiny{} \textbf{0} \textpm 4 } & {\tiny{} -4 \textpm 4 } & {\tiny{} -14 \textpm 4 } & {\tiny{} -3 \textpm 4 } & {\tiny{} 11 \textpm 4 }\tabularnewline
\hline 
{\tiny classic Expectiminimax with Completion} & {\tiny{} -18 \textpm 4 } & {\tiny{} -3 \textpm 4} & {\tiny{} \textbf{5} \textpm 3 } & {\tiny{} \textbf{0} \textpm 4 } & {\tiny{} -4 \textpm 4} & {\tiny{} -1 \textpm 4} & {\tiny{} \textbf{0} \textpm 4 } & {\tiny{} -10 \textpm 4}\tabularnewline
\hline 
{\tiny classic Expectiminimax} & {\tiny -16 \textpm 4 } & {\tiny{} \textbf{1} \textpm 4 } & {\tiny{} 0 \textpm 3 } & {\tiny{} \textbf{-1} \textpm 4 } & {\tiny{} 0 \textpm 4 } & {\tiny{} -2 \textpm 4 } & {\tiny{} \textbf{2} \textpm 4 } & {\tiny{} -25 \textpm 4}\tabularnewline
\hline 
{\tiny$\mctsh{}$ with Completion} & {\tiny -22 \textpm 4 } & {\tiny{} -22 \textpm 4 } & {\tiny{} -1 \textpm 3} & {\tiny{} -7 \textpm 4} & {\tiny{} -5 \textpm 4} & {\tiny{} -3 \textpm 4} & {\tiny{} \textbf{-1} \textpm 4} & {\tiny{} -21 \textpm 4}\tabularnewline
\hline 
{\tiny$\mctsh{}$} & {\tiny -20 \textpm 4 } & {\tiny{} -20 \textpm 4 } & {\tiny{} \textbf{5} \textpm 3 } & {\tiny{} -4 \textpm 4} & {\tiny{} -5 \textpm 4} & {\tiny{} -3 \textpm 4} & {\tiny{} \textbf{-1} \textpm 4} & {\tiny{} -40 \textpm 3}\tabularnewline
\hline 
{\scriptsize$\mctsh{'}$} & {\tiny -23 \textpm 4 } & {\tiny{} -26 \textpm 4 } & {\tiny{} \textbf{4} \textpm 3 } & {\tiny{} -5 \textpm 4} & {\tiny{} -5 \textpm 4} & {\tiny{} -8 \textpm 4} & {\tiny{} -5 \textpm 4} & {\tiny{} -65 \textpm 3}\tabularnewline
\hline 
{\scriptsize$\mctsh{\text{open}}$} & {\tiny -19 \textpm 4 } & {\tiny{} -26 \textpm 4 } & {\tiny{} \textbf{3} \textpm 3 } & {\tiny{} -5 \textpm 4} & {\tiny{} -18 \textpm 4 } & {\tiny{} -12 \textpm 4 } & {\tiny{} -38 \textpm 4 } & {\tiny{} -81 \textpm 2}\tabularnewline
\hline 
{\tiny$\mcts{}$} & {\tiny -100 \textpm{} 0 } & {\tiny{} -95 \textpm{} 1 } & {\tiny{} -19 \textpm{} 4 } & {\tiny -16 \textpm{} 4 } & {\tiny{} -71 \textpm{} 3 } & {\tiny{} -98 \textpm{} 1 } & {\tiny{} -66 \textpm{} 3 } & {\tiny{} -98 \textpm{} 1 }\tabularnewline
\hline 
\end{tabular}}{\tiny}{\tiny\par}
\par\end{centering}
\caption{Binary scores for all games of the evaluated algorithms without batching
against Expectiminimax with batching (All algorithms have 1s per action).}\label{tab:res_sans-batch-details-1}
\end{table*}

\subsection{Tuning of Search Algorithms}\label{subsec:tuning}

In this section, we present the tuning of parameterized search algorithms.

Tuning for the constant $C$ of $\mcts{}$, $\mctsh{\text{open}}$,
$\mctshd{}$ and $\mctsh{}$ for a search time if $10$ second are
described in Table~\ref{tab:tuning_10s_mean},  Table~\ref{tab:tab:tuning_10s_details},
Table~\ref{tab:tuning_10s_mean-new}, and Table~\ref{tab:tab:tuning_10s_details-new}.

Tuning for the constant $C$ of $\mcts{}$, $\mctsh{\text{open}}$,
$\mctshd{}$ and $\mctsh{}$ for a search time if $1$ second are
described in Table~\ref{tab:tuning_1s_mean},  Table~\ref{tab:tab:tuning_1s_details}, Table~\ref{tab:tuning_1s_mean-new}, and
Table~\ref{tab:tab:tuning_1s_details-new}.

We took the best parameters for each game and relaunched the article's
experience for each of them on the game in question.
\begin{table}
\begin{centering}
{\small{}%
\begin{tabular}{|c|c|c|c|}
\hline 
{\small$\mctsh{}$} & {\small mean} & {\small lower bound} & {\small upper bound}\tabularnewline
\hline 
\hline 
$2\sqrt{2}$ & -29.38 & -30.44 & -28.33 \tabularnewline
\hline 
$\frac{\sqrt{2}}{2}$ & {\small -27.21} & {\small -28.26} & {\small -26.14}\tabularnewline
\hline 
$\frac{\sqrt{2}}{4}$ & {\small -24.87} & {\small -25.91} & {\small -23.81}\tabularnewline
\hline 
$\frac{\sqrt{2}}{8}$ & {\small -19.95} & {\small -21.03} & {\small -18.85}\tabularnewline
\hline 
$\frac{\sqrt{2}}{16}$ & {\small -13.44} & {\small -14.57,} & {\small -12.29}\tabularnewline
\hline 
$\frac{\sqrt{2}}{32}$ & {\small -8.62} & {\small -9.78} & {\small -7.49}\tabularnewline
\hline 
$\frac{\sqrt{2}}{64}$ & {\small -11.96} & {\small -13.08} & {\small -10.85}\tabularnewline
\hline 
$\frac{\sqrt{2}}{128}$ & {\small -17.41} & {\small -18.52} & {\small -16.32}\tabularnewline
\hline 
\hline 
{\small$\mcts{}$} & {\small mean} & {\small lower bound} & {\small upper bound}\tabularnewline
\hline 
\hline 
$2\sqrt{2}$ & {\small -58.88} & {\small -59.73} & {\small -58.02}\tabularnewline
\hline 
$\sqrt{2}$ & {\small -59.7} & {\small -60.55} & {\small -58.86}\tabularnewline
\hline 
$\frac{\sqrt{2}}{2}$ & {\small -56.68} & {\small -57.53} & {\small -55.8}\tabularnewline
\hline 
$\frac{\sqrt{2}}{4}$ & {\small -58.81} & {\small -59.67} & {\small -57.95}\tabularnewline
\hline 
$\frac{\sqrt{2}}{8}$ & {\small -63.9} & {\small -64.71} & {\small -63.08}\tabularnewline
\hline 
$\frac{\sqrt{2}}{16}$ & {\small -69.78} & {\small -70.56} & {\small -69}\tabularnewline
\hline 
\end{tabular}}{\small\par}
\par\end{centering}
\caption{Average binary scores over all games of {\small$\protect\mctsh{}$
and $\protect\mcts{}$ for different values of $C$} against classic
Expectiminimax with batching (10 seconds per search).}\label{tab:tuning_10s_mean}
\end{table}
\begin{table*}[t]
\begin{centering}
{\tiny{}%
\begin{tabular}{|c|c|c|c|c|c|c|c|c|}
\hline 
{\small$\mctsh{}$} & {\tiny Azul} & {\tiny Backgammon} & {\tiny Can't Stop} & {\tiny Einstein würfelt nicht!} & {\tiny Kingdomino} & {\tiny Mighty Duel Kingdomino} & {\tiny Martian Dice} & {\tiny Shogun}\tabularnewline
\hline 
\hline 
$2\sqrt{2}$ & {\tiny -83 \textpm 2 } & {\tiny{} -23 \textpm 4 } & {\tiny{} -14 \textpm 4 } & {\tiny{} -3 \textpm 4 } & {\tiny{} -3 \textpm 4 } & {\tiny{} -6 \textpm 4 } & {\tiny{} -3 \textpm 4 } & {\tiny{} -99 \textpm 1}\tabularnewline
\hline 
$\frac{\sqrt{2}}{2}$ & {\tiny -71 \textpm 3 } & {\tiny{} -23 \textpm 4 } & {\tiny{} -6 \textpm 4 } & {\tiny{} -5 \textpm 4 } & {\tiny{} -9 \textpm 4 } & {\tiny{} -2 \textpm 4 } & {\tiny{} -4 \textpm 4 } & {\tiny{} -98 \textpm 1}\tabularnewline
\hline 
$\frac{\sqrt{2}}{4}$ & {\tiny -62 \textpm 3 } & {\tiny{} -22 \textpm 4 } & {\tiny{} -2 \textpm 3 } & {\tiny{} 1 \textpm 4 } & {\tiny{} -6 \textpm 4 } & {\tiny{} -7 \textpm 4 } & {\tiny{} -4 \textpm 4 } & {\tiny{} -97 \textpm 1}\tabularnewline
\hline 
$\frac{\sqrt{2}}{8}$ & {\tiny -33 \textpm 4 } & {\tiny{} -24 \textpm 4 } & {\tiny{} 4 \textpm 3 } & {\tiny{} -1 \textpm 4 } & {\tiny{} -6 \textpm 4 } & {\tiny{} -4 \textpm 4 } & {\tiny{} -4 \textpm 4 } & {\tiny{} -93 \textpm 1}\tabularnewline
\hline 
$\frac{\sqrt{2}}{16}$ & {\tiny 1 \textpm 4 } & {\tiny{} -21 \textpm 4 } & {\tiny{} 7 \textpm 3 } & {\tiny{} -5 \textpm 4 } & {\tiny{} -6 \textpm 4 } & {\tiny{} -3 \textpm 4 } & {\tiny{} -4 \textpm 4 } & {\tiny{} -77 \textpm 3}\tabularnewline
\hline 
$\frac{\sqrt{2}}{32}$ & {\tiny 24 \textpm 4 } & {\tiny{} -14 \textpm 4 } & {\tiny{} 6 \textpm 3 } & {\tiny{} -7 \textpm 4 } & {\tiny{} -6 \textpm 4 } & {\tiny{} 0 \textpm 4 } & {\tiny{} -5 \textpm 4 } & {\tiny{} -66 \textpm 3}\tabularnewline
\hline 
$\frac{\sqrt{2}}{64}$ & {\tiny 3 \textpm 4 } & {\tiny{} -13 \textpm 4 } & {\tiny{} 6 \textpm 3 } & {\tiny{} -14 \textpm 4 } & {\tiny{} -6 \textpm 4 } & {\tiny{} -2 \textpm 4 } & {\tiny{} -7 \textpm 4 } & {\tiny{} -63 \textpm 3}\tabularnewline
\hline 
$\frac{\sqrt{2}}{128}$ & {\tiny -51 \textpm 3 } & {\tiny{} -9 \textpm 4 } & {\tiny{} 3 \textpm 3 } & {\tiny{} -8 \textpm 4 } & {\tiny{} -5 \textpm 4 } & {\tiny{} 2 \textpm 4 } & {\tiny{} -8 \textpm 4 } & {\tiny{} -62 \textpm 3}\tabularnewline
\hline 
\hline 
{\small$\mcts{}$} & {\tiny Azul} & {\tiny Backgammon} & {\tiny Can't Stop} & {\tiny Einstein würfelt nicht!} & {\tiny Kingdomino} & {\tiny Mighty Duel Kingdomino} & {\tiny Martian Dice} & {\tiny Shogun}\tabularnewline
\hline 
\hline 
$2\sqrt{2}$ & {\tiny -99 \textpm 1} & {\tiny -95 \textpm 1} & {\tiny -7 \textpm 4} & {\tiny -12 \textpm 4} & {\tiny -35 \textpm 4} & {\tiny -85 \textpm 2} & {\tiny -40 \textpm 4} & {\tiny -98 \textpm 1}\tabularnewline
\hline 
$\sqrt{2}$ & {\tiny -99 \textpm 1} & {\tiny -95 \textpm 1} & {\tiny -12 \textpm 4} & {\tiny -12 \textpm 4} & {\tiny -34 \textpm 4} & {\tiny -84 \textpm 2} & {\tiny -45 \textpm 4} & {\tiny -98 \textpm 1}\tabularnewline
\hline 
$\frac{\sqrt{2}}{2}$ & {\tiny -97 \textpm 1} & {\tiny -94 \textpm 1} & {\tiny -9 \textpm 4} & {\tiny -10 \textpm 4} & {\tiny -23 \textpm 4} & {\tiny -78 \textpm 3} & {\tiny -45 \textpm 4} & {\tiny -98 \textpm 1}\tabularnewline
\hline 
$\frac{\sqrt{2}}{4}$ & {\tiny -98 \textpm 1} & {\tiny -95 \textpm 1} & {\tiny -12 \textpm 4} & {\tiny -13 \textpm 4} & {\tiny -24 \textpm 4} & {\tiny -79 \textpm 3} & {\tiny -52 \textpm 3} & {\tiny -97 \textpm 1}\tabularnewline
\hline 
$\frac{\sqrt{2}}{8}$ & {\tiny -100 \textpm 0} & {\tiny -95 \textpm 1} & {\tiny -11 \textpm 4} & {\tiny -28 \textpm 4} & {\tiny -34 \textpm 4} & {\tiny -81 \textpm 2} & {\tiny -65 \textpm 3} & {\tiny -98 \textpm 1}\tabularnewline
\hline 
$\frac{\sqrt{2}}{16}$ & {\tiny -100 \textpm 0} & {\tiny -95 \textpm 1} & {\tiny -20 \textpm 4} & {\tiny -48 \textpm 4} & {\tiny -42 \textpm 4} & {\tiny -82 \textpm 2} & {\tiny -73 \textpm 3} & {\tiny -98 \textpm 1}\tabularnewline
\hline 
\end{tabular}}{\tiny\par}
\par\end{centering}
\caption{Binary scores for all games of {\small$\protect\mctsh{}$ and $\protect\mcts{}$
for different values of $C$} against classic Expectiminimax with
batching (10 seconds per search).}\label{tab:tab:tuning_10s_details}
\end{table*}
\begin{table}
\begin{centering}
{\small{}%
\begin{tabular}{|c|c|c|c|}
\hline 
{\small$\mctsh{'}$} & {\small mean} & {\small lower bound} & {\small upper bound}\tabularnewline
\hline 
\hline 
$\sqrt{2}$ & {\small -29.6} & {\small -30.67} & {\small -28.55}\tabularnewline
\hline 
$\frac{\sqrt{2}}{2}$ & {\small -27.49} & {\small -28.52} & {\small -26.42}\tabularnewline
\hline 
$\frac{\sqrt{2}}{4}$ & {\small -24.78} & {\small -25.84} & {\small -23.71}\tabularnewline
\hline 
$\frac{\sqrt{2}}{8}$ & {\small -20.46} & {\small -21.56,} & {\small -19.36}\tabularnewline
\hline 
$\frac{\sqrt{2}}{16}$ & {\small -15.59} & {\small -16.7} & {\small -14.48}\tabularnewline
\hline 
$\frac{\sqrt{2}}{32}$ & {\small -13.82} & {\small -14.94} & {\small -12.68}\tabularnewline
\hline 
$\frac{\sqrt{2}}{64}$ & {\small -13.33} & {\small -14.46} & {\small -12.2}\tabularnewline
\hline 
$\frac{\sqrt{2}}{128}$ & {\small -20.19} & {\small -21.32} & {\small -19.06}\tabularnewline
\hline 
\hline 
{\small$\mctsh{\text{open}}$} & {\small mean} & {\small lower bound} & {\small upper bound}\tabularnewline
\hline 
\hline 
$\sqrt{2}$ & {\small -38.69} & {\small -39.71} & {\small -37.67}\tabularnewline
\hline 
$\frac{\sqrt{2}}{2}$ & {\small -38.01} & {\small -39.05,} & {\small -36.96}\tabularnewline
\hline 
$\frac{\sqrt{2}}{4}$ & {\small -36.28} & {\small -37.33} & {\small -35.23}\tabularnewline
\hline 
$\frac{\sqrt{2}}{8}$ & {\small -34.85} & {\small -35.91} & {\small -33.79}\tabularnewline
\hline 
$\frac{\sqrt{2}}{16}$ & {\small -29.78} & {\small -30.84} & {\small -28.73 }\tabularnewline
\hline 
$\frac{\sqrt{2}}{32}$ & {\small -28.27} & {\small -29.36} & {\small -27.2}\tabularnewline
\hline 
$\frac{\sqrt{2}}{64}$ & {\small -25.89} & {\small -26.98} & {\small -24.82}\tabularnewline
\hline 
$\frac{\sqrt{2}}{128}$ & {\small -33.03} & {\small -34.09} & {\small -31.95}\tabularnewline
\hline 
\end{tabular}}{\small\par}
\par\end{centering}
\caption{Average binary scores over all games of {\small$\protect\mctsh{'}$
and $\protect\mctsh{\text{open}}$ for different values of $C$} against
classic Expectiminimax with batching (10 seconds per search).}\label{tab:tuning_10s_mean-new}
\end{table}
\begin{table*}[t]
\begin{centering}
{\tiny{}%
\begin{tabular}{|c|c|c|c|c|c|c|c|c|}
\hline 
{\small$\mctsh{'}$} & {\tiny Azul} & {\tiny Backgammon} & {\tiny Can't Stop} & {\tiny Einstein würfelt nicht!} & {\tiny Kingdomino} & {\tiny Mighty Duel Kingdomino} & {\tiny Martian Dice} & {\tiny Shogun}\tabularnewline
\hline 
\hline 
$\sqrt{2}$ & {\tiny -78 \textpm 3 } & {\tiny{} -26 \textpm 4 } & {\tiny{} -13 \textpm 4 } & {\tiny{} -5 \textpm 4 } & {\tiny{} -5 \textpm 4 } & {\tiny{} -5 \textpm 4 } & {\tiny{} -8 \textpm 4 } & {\tiny{} -99 \textpm 1}\tabularnewline
\hline 
$\frac{\sqrt{2}}{2}$ & {\tiny -71 \textpm 3 } & {\tiny{} -26 \textpm 4 } & {\tiny{} -8 \textpm 4 } & {\tiny{} 0 \textpm 4 } & {\tiny{} -8 \textpm 4 } & {\tiny{} -5 \textpm 4 } & {\tiny{} -3 \textpm 4 } & {\tiny{} -99 \textpm 1}\tabularnewline
\hline 
$\frac{\sqrt{2}}{4}$ & {\tiny -57 \textpm 3 } & {\tiny{} -26 \textpm 4 } & {\tiny{} -2 \textpm 4 } & {\tiny{} 1 \textpm 4 } & {\tiny{} -9 \textpm 4 } & {\tiny{} -7 \textpm 4 } & {\tiny{} 0 \textpm 4 } & {\tiny{} -99 \textpm 1}\tabularnewline
\hline 
$\frac{\sqrt{2}}{8}$ & {\tiny -31 \textpm 4 } & {\tiny{} -23 \textpm 4 } & {\tiny{} 3 \textpm 3 } & {\tiny{} -3 \textpm 4 } & {\tiny{} -7 \textpm 4 } & {\tiny{} -5 \textpm 4 } & {\tiny{} -2 \textpm 4 } & {\tiny{} -96 \textpm 1}\tabularnewline
\hline 
$\frac{\sqrt{2}}{16}$ & {\tiny 6 \textpm 4 } & {\tiny{} -20 \textpm 4 } & {\tiny{} 5 \textpm 3 } & {\tiny{} -5 \textpm 4 } & {\tiny{} -8 \textpm 4 } & {\tiny{} -6 \textpm 4 } & {\tiny{} -6 \textpm 4 } & {\tiny{} -91 \textpm 2}\tabularnewline
\hline 
$\frac{\sqrt{2}}{32}$ & {\tiny 20 \textpm 4 } & {\tiny{} -20 \textpm 4 } & {\tiny{} 2 \textpm 3 } & {\tiny{} -11 \textpm 4 } & {\tiny{} -10 \textpm 4 } & {\tiny{} -2 \textpm 4 } & {\tiny{} -6 \textpm 4 } & {\tiny{} -83 \textpm 2}\tabularnewline
\hline 
$\frac{\sqrt{2}}{64}$ & {\tiny 8 \textpm 4 } & {\tiny{} -14 \textpm 4 } & {\tiny{} 6 \textpm 3 } & {\tiny{} -13 \textpm 4 } & {\tiny{} -6 \textpm 4 } & {\tiny{} 0 \textpm 4 } & {\tiny{} -9 \textpm 4 } & {\tiny{} -77 \textpm 2}\tabularnewline
\hline 
$\frac{\sqrt{2}}{128}$ & {\tiny -47 \textpm 4 } & {\tiny{} -16 \textpm 4 } & {\tiny{} 7 \textpm 3 } & {\tiny{} -15 \textpm 4 } & {\tiny{} -9 \textpm 4 } & {\tiny{} 4 \textpm 4 } & {\tiny{} -16 \textpm 4 } & {\tiny{} -71 \textpm 3}\tabularnewline
\hline 
\hline 
{\small$\mctsh{\text{open}}$} & {\tiny Azul} & {\tiny Backgammon} & {\tiny Can't Stop} & {\tiny Einstein würfelt nicht!} & {\tiny Kingdomino} & {\tiny Mighty Duel Kingdomino} & {\tiny Martian Dice} & {\tiny Shogun}\tabularnewline
\hline 
\hline 
$\sqrt{2}$ & {\tiny -81 \textpm 2 } & {\tiny{} -38 \textpm 4 } & {\tiny{} -23 \textpm 3 } & {\tiny{} -3 \textpm 4 } & {\tiny{} -24 \textpm 4 } & {\tiny{} -21 \textpm 4 } & {\tiny{} -21 \textpm 4 } & {\tiny{} -98 \textpm 1}\tabularnewline
\hline 
$\frac{\sqrt{2}}{2}$ & {\tiny -75 \textpm 3 } & {\tiny{} -34 \textpm 4 } & {\tiny{} -28 \textpm 3 } & {\tiny{} 0 \textpm 4 } & {\tiny{} -26 \textpm 4 } & {\tiny{} -21 \textpm 4 } & {\tiny{} -22 \textpm 4 } & {\tiny{} -98 \textpm 1}\tabularnewline
\hline 
$\frac{\sqrt{2}}{4}$ & {\tiny -59 \textpm 3 } & {\tiny{} -37 \textpm 4 } & {\tiny{} -19 \textpm 3 } & {\tiny{} 2 \textpm 4 } & {\tiny{} -27 \textpm 4 } & {\tiny{} -19 \textpm 4 } & {\tiny{} -34 \textpm 4 } & {\tiny{} -98 \textpm 1}\tabularnewline
\hline 
$\frac{\sqrt{2}}{8}$ & {\tiny -33 \textpm 4 } & {\tiny{} -39 \textpm 4 } & {\tiny{} -18 \textpm 3 } & {\tiny{} -3 \textpm 4 } & {\tiny{} -29 \textpm 4 } & {\tiny{} -24 \textpm 4 } & {\tiny{} -36 \textpm 4 } & {\tiny{} -97 \textpm 1}\tabularnewline
\hline 
$\frac{\sqrt{2}}{16}$ & {\tiny 0 \textpm 4 } & {\tiny{} -36 \textpm 4 } & {\tiny{} -9 \textpm 3 } & {\tiny{} -3 \textpm 4 } & {\tiny{} -29 \textpm 4 } & {\tiny{} -23 \textpm 4 } & {\tiny{} -43 \textpm 4 } & {\tiny{} -94 \textpm 1}\tabularnewline
\hline 
$\frac{\sqrt{2}}{32}$ & {\tiny 7 \textpm 4 } & {\tiny{} -33 \textpm 4 } & {\tiny{} -4 \textpm 3 } & {\tiny{} -10 \textpm 4 } & {\tiny{} -27 \textpm 4 } & {\tiny{} -21 \textpm 4 } & {\tiny{} -45 \textpm 4 } & {\tiny{} -94 \textpm 1}\tabularnewline
\hline 
$\frac{\sqrt{2}}{64}$ & {\tiny 1 \textpm 4 } & {\tiny{} -27 \textpm 4 } & {\tiny{} 2 \textpm 3 } & {\tiny{} -11 \textpm 4 } & {\tiny{} -27 \textpm 4 } & {\tiny{} -19 \textpm 4 } & {\tiny{} -37 \textpm 4 } & {\tiny{} -89 \textpm 2}\tabularnewline
\hline 
$\frac{\sqrt{2}}{128}$ & {\tiny -51 \textpm 3 } & {\tiny{} -23 \textpm 4 } & {\tiny{} 4 \textpm 3 } & {\tiny{} -17 \textpm 4 } & {\tiny{} -35 \textpm 4 } & {\tiny{} -19 \textpm 4 } & {\tiny{} -41 \textpm 4 } & {\tiny{} -83 \textpm 2}\tabularnewline
\hline 
\end{tabular}}{\tiny\par}
\par\end{centering}
\caption{Binary scores for all games of {\small$\protect\mctsh{'}$ and $\protect\mctsh{\text{open}}$
for different values of $C$} against classic Expectiminimax with
batching (10 seconds per search).}\label{tab:tab:tuning_10s_details-new}
\end{table*}

\begin{table}
\begin{centering}
{\small{}%
\begin{tabular}{|c|c|c|c|}
\hline 
{\small$\mctsh{}$} & {\small mean} & {\small lower bound} & {\small upper bound}\tabularnewline
\hline 
\hline 
$2\sqrt{2}$ & {\small -30.77} & {\small -31.77} & {\small -29.74}\tabularnewline
\hline 
$\frac{\sqrt{2}}{2}$ & {\small -29.06} & {\small -30.09} & {\small -28.03}\tabularnewline
\hline 
$\frac{\sqrt{2}}{4}$ & {\small -29.11} & {\small -30.11} & {\small -28.07}\tabularnewline
\hline 
$\frac{\sqrt{2}}{8}$ & {\small -26.16} & {\small -27.21} & {\small -25.1}\tabularnewline
\hline 
$\frac{\sqrt{2}}{16}$ & {\small -20.65} & {\small -21.72} & {\small -19.55}\tabularnewline
\hline 
$\frac{\sqrt{2}}{32}$ & {\small -16.45} & {\small -17.58} & {\small -15.32}\tabularnewline
\hline 
$\frac{\sqrt{2}}{64}$ & {\small -17.81} & {\small -18.95} & {\small -16.67}\tabularnewline
\hline 
$\frac{\sqrt{2}}{128}$ & {\small -19.8} & {\small -20.89} & {\small -18.69}\tabularnewline
\hline 
\hline 
{\small$\mcts{}$} & {\small mean} & {\small lower bound} & {\small upper bound}\tabularnewline
\hline 
\hline 
$2\sqrt{2}$ & {\small -71.66} & {\small -72.38} & {\small -70.95}\tabularnewline
\hline 
$\sqrt{2}$ & {\small -71.66} & {\small -72.38} & {\small -70.96}\tabularnewline
\hline 
$\frac{\sqrt{2}}{2}$ & {\small -71.13} & {\small -71.85} & {\small -70.4}\tabularnewline
\hline 
$\frac{\sqrt{2}}{4}$ & {\small -73.18} & {\small -73.89} & {\small -72.47}\tabularnewline
\hline 
$\frac{\sqrt{2}}{8}$ & {\small -75.36} & {\small -76.05} & {\small -74.66}\tabularnewline
\hline 
$\frac{\sqrt{2}}{16}$ & {\small -77.87} & {\small -78.53} & {\small -77.22}\tabularnewline
\hline 
\end{tabular}}{\small\par}
\par\end{centering}
\caption{Average binary scores over all games of {\small$\protect\mctsh{}$
and $\protect\mcts{}$ for different values of $C$} against classic
Expectiminimax with batching (1 seconds per search).}\label{tab:tuning_1s_mean}
\end{table}
\begin{table*}[t]
\begin{centering}
{\tiny{}%
\begin{tabular}{|c|c|c|c|c|c|c|c|c|}
\hline 
{\small$\mctsh{}$} & {\tiny Azul} & {\tiny Backgammon} & {\tiny Can't Stop} & {\tiny Einstein würfelt nicht!} & {\tiny Kingdomino} & {\tiny Mighty Duel Kingdomino} & {\tiny Martian Dice} & {\tiny Shogun}\tabularnewline
\hline 
\hline 
$2\sqrt{2}$ & {\tiny -89 \textpm 2 } & {\tiny{} -35 \textpm 4 } & {\tiny{} 5 \textpm 3 } & {\tiny{} -6 \textpm 4 } & {\tiny{} -10 \textpm 4 } & {\tiny{} -9 \textpm 4 } & {\tiny{} -5 \textpm 4 } & {\tiny{} -98 \textpm 1}\tabularnewline
\hline 
$\frac{\sqrt{2}}{2}$ & {\tiny -87 \textpm 2 } & {\tiny{} -35 \textpm 4 } & {\tiny{} 5 \textpm 3 } & {\tiny{} -5 \textpm 4 } & {\tiny{} -4 \textpm 4 } & {\tiny{} -7 \textpm 4 } & {\tiny{} -2 \textpm 4 } & {\tiny{} -97 \textpm 1}\tabularnewline
\hline 
$\frac{\sqrt{2}}{4}$ & {\tiny -83 \textpm 2 } & {\tiny{} -31 \textpm 4 } & {\tiny{} 2 \textpm 3 } & {\tiny{} -4 \textpm 4 } & {\tiny{} -6 \textpm 4 } & {\tiny{} -11 \textpm 4 } & {\tiny{} -2 \textpm 4 } & {\tiny{} -97 \textpm 1}\tabularnewline
\hline 
$\frac{\sqrt{2}}{8}$ & {\tiny -72 \textpm 3 } & {\tiny{} -29 \textpm 4 } & {\tiny{} 1 \textpm 3 } & {\tiny{} -2 \textpm 4 } & {\tiny{} -10 \textpm 4 } & {\tiny{} -7 \textpm 4 } & {\tiny{} 1 \textpm 4 } & {\tiny{} -93 \textpm 2}\tabularnewline
\hline 
$\frac{\sqrt{2}}{16}$ & {\tiny -46 \textpm 4 } & {\tiny{} -25 \textpm 4 } & {\tiny{} 3 \textpm 3 } & {\tiny{} -4 \textpm 4 } & {\tiny{} -5 \textpm 4 } & {\tiny{} -6 \textpm 4 } & {\tiny{} -1 \textpm 4 } & {\tiny{} -81 \textpm 2}\tabularnewline
\hline 
$\frac{\sqrt{2}}{32}$ & {\tiny -24 \textpm 4 } & {\tiny{} -23 \textpm 4 } & {\tiny{} 5 \textpm 3 } & {\tiny{} -10 \textpm 4 } & {\tiny{} 0 \textpm 4 } & {\tiny{} -8 \textpm 4 } & {\tiny{} -4 \textpm 4 } & {\tiny{} -67 \textpm 3}\tabularnewline
\hline 
$\frac{\sqrt{2}}{64}$ & {\tiny -49 \textpm 4 } & {\tiny{} -23 \textpm 4 } & {\tiny{} 3 \textpm 3 } & {\tiny{} -11 \textpm 4 } & {\tiny{} -6 \textpm 4 } & {\tiny{} -5 \textpm 4 } & {\tiny{} -4 \textpm 4 } & {\tiny{} -47 \textpm 3}\tabularnewline
\hline 
$\frac{\sqrt{2}}{128}$ & {\tiny -78 \textpm 3 } & {\tiny{} -19 \textpm 4 } & {\tiny{} 3 \textpm 3 } & {\tiny{} -8 \textpm 4 } & {\tiny{} -2 \textpm 4 } & {\tiny{} -5 \textpm 4 } & {\tiny{} -9 \textpm 4 } & {\tiny{} -41 \textpm 3}\tabularnewline
\hline 
\hline 
{\small$\mcts{}$} & {\tiny Azul} & {\tiny Backgammon} & {\tiny Can't Stop} & {\tiny Einstein würfelt nicht!} & {\tiny Kingdomino} & {\tiny Mighty Duel Kingdomino} & {\tiny Martian Dice} & {\tiny Shogun}\tabularnewline
\hline 
\hline 
$2\sqrt{2}$ & {\tiny -100 \textpm 0 } & {\tiny{} -95 \textpm 1 } & {\tiny{} -20 \textpm 4 } & {\tiny{} -19 \textpm 4 } & {\tiny{} -78 \textpm 3 } & {\tiny{} -98 \textpm 1 } & {\tiny{} -66 \textpm 3 } & {\tiny{} -99 \textpm 0}\tabularnewline
\hline 
$\sqrt{2}$ & {\tiny -100 \textpm 0 } & {\tiny{} -95 \textpm 1 } & {\tiny{} -22 \textpm 4 } & {\tiny{} -16 \textpm 4 } & {\tiny{} -77 \textpm 3 } & {\tiny{} -98 \textpm 1 } & {\tiny{} -67 \textpm 3 } & {\tiny{} -99 \textpm 0}\tabularnewline
\hline 
$\frac{\sqrt{2}}{2}$ & {\tiny -100 \textpm 0 } & {\tiny{} -95 \textpm 1 } & {\tiny{} -19 \textpm 4 } & {\tiny{} -18 \textpm 4 } & {\tiny{} -71 \textpm 3 } & {\tiny{} -98 \textpm 1 } & {\tiny{} -70 \textpm 3 } & {\tiny{} -99 \textpm 0}\tabularnewline
\hline 
$\frac{\sqrt{2}}{4}$ & {\tiny -100 \textpm 0 } & {\tiny{} -96 \textpm 1 } & {\tiny{} -20 \textpm 4 } & {\tiny{} -26 \textpm 4 } & {\tiny{} -71 \textpm 3 } & {\tiny{} -98 \textpm 1 } & {\tiny{} -74 \textpm 3 } & {\tiny{} -99 \textpm 0}\tabularnewline
\hline 
$\frac{\sqrt{2}}{8}$ & {\tiny -100 \textpm 0 } & {\tiny{} -96 \textpm 1 } & {\tiny{} -26 \textpm 4 } & {\tiny{} -40 \textpm 4 } & {\tiny{} -73 \textpm 3 } & {\tiny{} -98 \textpm 1 } & {\tiny{} -72 \textpm 3 } & {\tiny{} -98 \textpm 1}\tabularnewline
\hline 
$\frac{\sqrt{2}}{16}$ & {\tiny -100 \textpm 0 } & {\tiny{} -95 \textpm 1 } & {\tiny{} -23 \textpm 3 } & {\tiny{} -54 \textpm 3 } & {\tiny{} -78 \textpm 3 } & {\tiny{} -98 \textpm 1 } & {\tiny{} -76 \textpm 3 } & {\tiny{} -99 \textpm 0}\tabularnewline
\hline 
\end{tabular}}{\tiny\par}
\par\end{centering}
\caption{Binary scores for all games of {\small$\protect\mctsh{}$ and $\protect\mcts{}$
for different values of $C$} against classic Expectiminimax with
batching (1 seconds per search).}\label{tab:tab:tuning_1s_details}
\end{table*}

\begin{table*}
\begin{centering}
{\small{}%
\begin{tabular}{|c|c|c|c|}
\hline 
{\small$\mctsh{'}$} & {\small mean} & {\small lower bound} & {\small upper bound}\tabularnewline
\hline 
\hline 
$\sqrt{2}$ & {\small -30.07} & {\small -31.07} & {\small -29.07}\tabularnewline
\hline 
$\frac{\sqrt{2}}{2}$ & {\small -29.12} & {\small -30.14} & {\small -28.1}\tabularnewline
\hline 
$\frac{\sqrt{2}}{4}$ & {\small -28.87} & {\small -29.88} & {\small -27.84}\tabularnewline
\hline 
$\frac{\sqrt{2}}{8}$ & {\small -26.31} & {\small -27.36} & {\small -25.25}\tabularnewline
\hline 
$\frac{\sqrt{2}}{16}$ & {\small -22.93} & {\small -24} & {\small -21.84}\tabularnewline
\hline 
$\frac{\sqrt{2}}{32}$ & {\small -19.34} & {\small -20.44} & {\small -18.22}\tabularnewline
\hline 
$\frac{\sqrt{2}}{64}$ & {\small -22.08} & {\small -23.18} & {\small -20.97}\tabularnewline
\hline 
$\frac{\sqrt{2}}{128}$ & {\small -26.12} & {\small -27.2} & {\small -25.03}\tabularnewline
\hline 
\hline 
{\small$\mctsh{\text{open}}$} & {\small mean} & {\small lower bound} & {\small upper bound}\tabularnewline
\hline 
\hline 
$\sqrt{2}$ & {\small -37.01} & {\small -38.04} & {\small -36}\tabularnewline
\hline 
$\frac{\sqrt{2}}{2}$ & {\small -38.06} & {\small -39.07} & {\small -37.05}\tabularnewline
\hline 
$\frac{\sqrt{2}}{4}$ & {\small -36.8} & {\small -37.82} & {\small -35.77}\tabularnewline
\hline 
$\frac{\sqrt{2}}{8}$ & {\small -35.73} & {\small -36.75} & {\small -34.71}\tabularnewline
\hline 
$\frac{\sqrt{2}}{16}$ & {\small -32.7} & {\small -33.74} & {\small -31.65}\tabularnewline
\hline 
$\frac{\sqrt{2}}{32}$ & {\small -30.04 } & {\small -31.11} & {\small -28.99}\tabularnewline
\hline 
$\frac{\sqrt{2}}{64}$ & {\small -30.15} & {\small -31.23} & {\small -29.07}\tabularnewline
\hline 
$\frac{\sqrt{2}}{128}$ & {\small -33.61} & {\small -34.66,} & {\small -32.56}\tabularnewline
\hline 
\end{tabular}}{\small\par}
\par\end{centering}
\caption{Average binary scores over all games of {\small$\protect\mctsh{'}$
and $\protect\mctsh{\text{open}}$ for different values of $C$} against
classic Expectiminimax with batching (1 seconds per search).}\label{tab:tuning_1s_mean-new}
\end{table*}
\begin{table*}[t]
\begin{centering}
{\tiny{}%
\begin{tabular}{|c|c|c|c|c|c|c|c|c|}
\hline 
{\small$\mctsh{'}$} & {\tiny Azul} & {\tiny Backgammon} & {\tiny Can't Stop} & {\tiny Einstein würfelt nicht!} & {\tiny Kingdomino} & {\tiny Mighty Duel Kingdomino} & {\tiny Martian Dice} & {\tiny Shogun}\tabularnewline
\hline 
\hline 
$\sqrt{2}$ & {\tiny -88 \textpm 2 } & {\tiny{} -43 \textpm 4 } & {\tiny{} 4 \textpm 3 } & {\tiny{} -1 \textpm 4 } & {\tiny{} -6 \textpm 4 } & {\tiny{} -3 \textpm 4 } & {\tiny{} -5 \textpm 4 } & {\tiny{} -99 \textpm 1}\tabularnewline
\hline 
$\frac{\sqrt{2}}{2}$ & {\tiny{} -85 \textpm 2 } & {\tiny{} -38 \textpm 4 } & {\tiny{} 4 \textpm 3 } & {\tiny{} 0 \textpm 4 } & {\tiny{} -3 \textpm 4 } & {\tiny{} -9 \textpm 4 } & {\tiny{} -2 \textpm 4 } & {\tiny{} -99 \textpm 0}\tabularnewline
\hline 
$\frac{\sqrt{2}}{4}$ & {\tiny{} -80 \textpm 2 } & {\tiny{} -42 \textpm 4 } & {\tiny{} 1 \textpm 3 } & {\tiny{} 0 \textpm 4 } & {\tiny -3 \textpm 4 } & {\tiny{} -7 \textpm 4 } & {\tiny{} -1 \textpm 4 } & {\tiny{} -99 \textpm 1}\tabularnewline
\hline 
$\frac{\sqrt{2}}{8}$ & {\tiny{} -68 \textpm 3 } & {\tiny{} -33 \textpm 4 } & {\tiny{} 2 \textpm 3 } & {\tiny{} -1 \textpm 4 } & {\tiny{} -4 \textpm 4 } & {\tiny{} -8 \textpm 4 } & {\tiny{} 0 \textpm 4 } & {\tiny{} -98 \textpm 1}\tabularnewline
\hline 
$\frac{\sqrt{2}}{16}$ & {\tiny{} -44 \textpm 4 } & {\tiny{} -34 \textpm 4 } & {\tiny{} 3 \textpm 3 } & {\tiny{} -3 \textpm 4 } & {\tiny{} -8 \textpm 4 } & {\tiny{} -5 \textpm 4 } & {\tiny{} 2 \textpm 4 } & {\tiny{} -93 \textpm 1}\tabularnewline
\hline 
$\frac{\sqrt{2}}{32}$ & {\tiny{} -16 \textpm 4 } & {\tiny{} -29 \textpm 4 } & {\tiny{} 4 \textpm 3 } & {\tiny{} -12 \textpm 4 } & {\tiny{} -4 \textpm 4 } & {\tiny{} -8 \textpm 4 } & {\tiny{} -7 \textpm 4 } & {\tiny{} -82 \textpm 2}\tabularnewline
\hline 
$\frac{\sqrt{2}}{64}$ & {\tiny{} -45 \textpm 4 } & {\tiny{} -23 \textpm 4 } & {\tiny{} 4 \textpm 3 } & {\tiny{} -14 \textpm 4 } & {\tiny{} -8 \textpm 4 } & {\tiny{} -8 \textpm 4 } & {\tiny{} -12 \textpm 4 } & {\tiny{} -71 \textpm 3}\tabularnewline
\hline 
$\frac{\sqrt{2}}{128}$ & {\tiny{} -77 \textpm 3 } & {\tiny{} -27 \textpm 4 } & {\tiny{} 3 \textpm 3 } & {\tiny{} -13 \textpm 4 } & {\tiny{} -5 \textpm 4 } & {\tiny{} -6 \textpm 4 } & {\tiny{} -14 \textpm 4 } & {\tiny{} -70 \textpm 3}\tabularnewline
\hline 
\hline 
{\small$\mctsh{\text{open}}$} & {\tiny Azul} & {\tiny Backgammon} & {\tiny Can't Stop} & {\tiny Einstein würfelt nicht!} & {\tiny Kingdomino} & {\tiny Mighty Duel Kingdomino} & {\tiny Martian Dice} & {\tiny Shogun}\tabularnewline
\hline 
\hline 
$\sqrt{2}$ & {\scriptsize -87 \textpm 2 } & {\scriptsize{} -44 \textpm 4 } & {\scriptsize{} -2 \textpm 3 } & {\scriptsize{} -6 \textpm 4 } & {\scriptsize{} -13 \textpm 4 } & {\scriptsize{} -14 \textpm 4 } & {\scriptsize{} -31 \textpm 4 } & {\scriptsize{} -99 \textpm 0}\tabularnewline
\hline 
$\frac{\sqrt{2}}{2}$ & {\scriptsize{} -85 \textpm 2 } & {\scriptsize{} -46 \textpm 4 } & {\scriptsize{} -1 \textpm 3 } & {\scriptsize{} -6 \textpm 4 } & {\scriptsize{} -17 \textpm 4 } & {\scriptsize{} -17 \textpm 4 } & {\scriptsize{} -33 \textpm 4 } & {\scriptsize{} -99 \textpm 0}\tabularnewline
\hline 
$\frac{\sqrt{2}}{4}$ & {\scriptsize{} -81 \textpm 2 } & {\scriptsize{} -43 \textpm 4 } & {\scriptsize{} 1 \textpm 3 } & {\scriptsize{} -6 \textpm 4 } & {\scriptsize{} -16 \textpm 4 } & {\scriptsize{} -16 \textpm 4 } & {\scriptsize{} -32 \textpm 4 } & {\scriptsize{} -99 \textpm 0}\tabularnewline
\hline 
$\frac{\sqrt{2}}{8}$ & {\scriptsize{} -69 \textpm 3 } & {\scriptsize{} -43 \textpm 4 } & {\scriptsize{} 2 \textpm 3 } & {\scriptsize{} -3 \textpm 4 } & {\scriptsize{} -18 \textpm 4 } & {\scriptsize{} -17 \textpm 4 } & {\scriptsize{} -39 \textpm 4 } & {\scriptsize{} -99 \textpm 0}\tabularnewline
\hline 
$\frac{\sqrt{2}}{16}$ & {\scriptsize{} -42 \textpm 4 } & {\scriptsize{} -36 \textpm 4 } & {\scriptsize{} 1 \textpm 3 } & {\scriptsize{} -5 \textpm 4 } & {\scriptsize{} -18 \textpm 4 } & {\scriptsize{} -18 \textpm 4 } & {\scriptsize{} -45 \textpm 4 } & {\scriptsize{} -98 \textpm 1}\tabularnewline
\hline 
$\frac{\sqrt{2}}{32}$ & {\scriptsize{} -22 \textpm 4 } & {\scriptsize{} -28 \textpm 4 } & {\scriptsize{} -1 \textpm 3 } & {\scriptsize{} -14 \textpm 4 } & {\scriptsize{} -21 \textpm 4 } & {\scriptsize{} -16 \textpm 4 } & {\scriptsize{} -46 \textpm 4 } & {\scriptsize{} -91 \textpm 2}\tabularnewline
\hline 
$\frac{\sqrt{2}}{64}$ & {\scriptsize{} -47 \textpm 4 } & {\scriptsize{} -28 \textpm 4 } & {\scriptsize{} 3 \textpm 3 } & {\scriptsize{} -10 \textpm 4 } & {\scriptsize{} -18 \textpm 4 } & {\scriptsize{} -13 \textpm 4 } & {\scriptsize{} -45 \textpm 4 } & {\scriptsize{} -83 \textpm 2}\tabularnewline
\hline 
$\frac{\sqrt{2}}{128}$ & {\scriptsize{} -80 \textpm 2 } & {\scriptsize{} -27 \textpm 4 } & {\scriptsize{} 6 \textpm 3 } & {\scriptsize{} -13 \textpm 4 } & {\scriptsize{} -20 \textpm 4 } & {\scriptsize{} -12 \textpm 4 } & {\scriptsize{} -44 \textpm 4 } & {\scriptsize{} -80 \textpm 2}\tabularnewline
\hline 
\end{tabular}}{\tiny\par}
\par\end{centering}
\caption{Binary scores for all games of {\small$\protect\mctsh{'}$ and $\protect\mctsh{\text{open}}$
for different values of $C$} against classic Expectiminimax with
batching (1 seconds per search).}\label{tab:tab:tuning_1s_details-new}
\end{table*}

\subsection{Games}\label{subsec:Games}

We present in this section the games on which we conduct our comparison experiments for search algorithms,
namely EinStein würfelt nicht!, Backgammon, Azul, Kingdomino, Mighty
Duel Kingdomino, Martian Dice, Shogun, and Can’t Stop. A comparison
of the different characteristics of these games, namely state space
estimation, proportion of chance estimation, and BoardGameGeek game
rating, is provided in Table~\ref{tab:Some-characteristics-of}.
\begin{table*}
\begin{centering}
\begin{tabular}{|c|c|c|c|}
\hline 
 & game tree size & chance weights & BGG rating\tabularnewline
\hline 
\hline 
EinStein würfelt nicht! & $10^{7}$ & $0.63$ & $6.7$\tabularnewline
\hline 
Backgammon & $10^{133}$ & $0.52$ & $6.6$\tabularnewline
\hline 
Azul & $10^{12}$ & $0.52$ & $7.7$\tabularnewline
\hline 
Kingdomino & $10^{14}$ & $0.45$ & $7.3$\tabularnewline
\hline 
Mighty Duel Kingdomino & $10^{24}$ & $0.48$ & $7.3$\tabularnewline
\hline 
Martian Dice & $10^{882}$ & $0.86$ & $6.1$\tabularnewline
\hline 
Can’t Stop & $10^{103}$ & $0.71$ & $6.9$\tabularnewline
\hline 
Shogun & $10^{32}$ & $0.28$ & $5.8$\tabularnewline
\hline 
\end{tabular}
\par\end{centering}
\caption{Some characteristics of the benchmark games: the game tree size estimation
is $b^{d}$ where $b$ is the mean branching factor and $d$ is the
mean number of states in a full match ; the combinatorial weight of
chance estimation is defined as the mean over matches $m$ of $\frac{\sum_{s\in m_{c}}\log b_{s}}{\sum_{s\in m_{c}}\log b_{s}+\sum_{s\in m_{d}}\log b_{s}}$
where $m_{c}$ is the set of chance states of $m$, $m_{d}$ is the
set of decision states of $m$, and $b_{s}$ is the branching factor
of state $s$. Values are computed from $10000$ random matchs. The BGG rating
is the game’s
rating as given by the BoardGameGeek community (\protect\url{https://boardgamegeek.com}).}\label{tab:Some-characteristics-of}
\end{table*}

\subsubsection{\emph{EinStein Würfelt Nicht!}}

We start by presenting the game \emph{EinStein würfelt nicht!. }

\begin{figure}
\centering{}\includegraphics[scale=0.15]{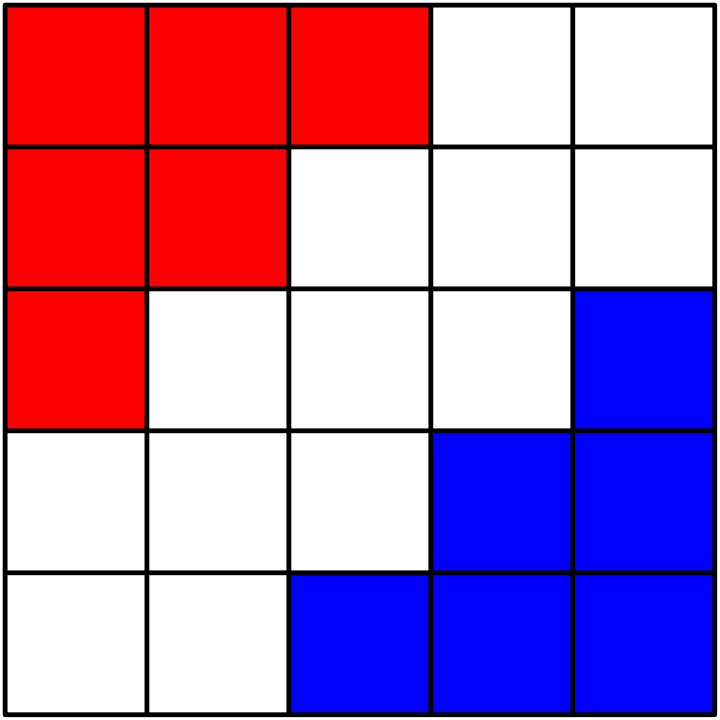}\caption{The two starting areas at EinStein wuerfelt nicht.}\label{einstein}
\end{figure}

This game is played on a $5\times5$ square board. The objective is
to reach with one of his pieces his arrival position which is the
square of the corner of the board located in the opponent's starting
area. The two starting areas are described in Figure \ref{einstein}.
The game has an initialization phase where the players will place
their 6 pieces on the board. Each player's pieces are numbered from
1 to 6. During the initialization phase, the pieces are arranged randomly
but symmetrically with respect to the games (central symmetry). Then
a race phase starts. As soon as a player reaches his arrival square
with one of his pieces, the game is over. At the start of each turn,
during the race phase, a 6-sided die is rolled. It indicates the current
player's piece that he can move (the one with the same number as the
result of the die). If there is no piece on the board corresponding
to the number on the die, the piece to be moved is the one whose number
is closest to the result of the die. If there are two possibilities,
the player chooses. A piece's possible move is one of three moves
to an adjacent square that bring the piece closer to that player's
arrival square (horizontal move, vertical move, or diagonal move).
If the piece ends its movement on another piece (no matter which player
it belongs to), the other piece is removed from the board. 

We use these rules in this paper (they are the Computer Olympiad rules).
However, variants do exist, with respect to the initialization phase,
where for example the first player places his pieces in his starting
area and then the second player places his pieces in his starting
area symmetrically (Polygames rules~\cite{cazenave2020polygames}).
In the Ludii \cite{Piette2020Ludii} implementation, during the initialization
phase, in turn, each player places one of his pieces in his starting
area. 

Despite its great simplicity and the element of luck, it is a game
that has tactics and strategies.

The more recent programs at \emph{EinStein würfelt nicht!} are based
on MCTS \cite{chen2023design,chu2017agent,lu2015einstein,lorentz2011mcts}.
There are also works on endgame tablebase \cite{turner2012einstein}
and on solving \emph{EinStein würfelt nicht!} \cite{bonnet2017toward}.

According to the BoardGameGeek community, EinStein würfelt nicht!
ranks as the 600th-best abstract game and 14,827th overall.

\subsubsection{Backgammon}

Backgammon is one of the oldest known board games, with origins estimated to date back to around 3000 BCE. Backgammon provides a long-standing benchmark for
evaluating decision-making under uncertainty in stochastic perfect-information
games~\cite{tesauro2002programming,van2007monte,pollack1997coevolution}. 

Backgammon is played on a board of 24 triangular points. Each player
starts with 15 checkers. The objective is to be the first player to
bear off all of one’s checkers from the board. On each turn, a player
rolls two dice and must move checkers according to the dice values.
A checker may move to an open point (unoccupied or occupied by the
player’s own checkers) or to a point occupied by exactly one opponent
checker, in which case the opponent checker is hit and placed on the
bar. Checkers on the bar must re-enter the board before any other
move can be played. Once all of a player’s checkers are in their home
board (the last six points), they may begin bearing off checkers according
to the dice rolls. Complete rules of Backgammon are available at boardgamegeek.com~\cite{bgg:backgammon}.

Historically, Backgammon played a central role in demonstrating the
effectiveness of temporal-difference learning with neural networks,
most notably through Tesauro’s TD-Gammon~\cite{tesauro1995temporal,tesauro2002programming},
which reached strong human-expert level using self-play and function
approximation. AlphaZero also reaches this success recently~\cite{antonoglou2021planning}.
Deciding whether a player can win at Backgammon is EXPTIME-Hard~\cite{witter2021backgammon}.

According to the BoardGameGeek community, Backgammon ranks as the
141st-best abstract game and 1,641st overall.

\subsubsection{Azul}

Azul is a game featuring \textit{chaining}, \textit{grid coverage},
\textit{open drafting}, \textit{pattern building}, and \textit{set
collection} mechanics. The core mechanics of Azul consist of drafting
5 groups of 4 randomly drawn colored tiles and placing them on a personal
board to maximize points under geometric placement constraints, while
managing penalties and strategic risk arising from indirect opponent
interaction and chance. Rules of Azul are available at boardgamegeek.com~\cite{bgg:azul}.

According to the BoardGameGeek community, Azul has a complexity comparable
to Backgammon, ranks as the second-best abstract game and 97th overall.
It has sold over two million copies worldwide, and won both the Spiel
des Jahres and the As d’Or (game of the year) in 2018, two of the
most prestigious awards in the board game industry.

The mechanics of Azul have been analyzed, showing that they promote
indirect and continuous sabotage through resource denial while maintaining
a balance between individual optimization and competitive player interaction
\cite{acatrinei2024tactical}. 

\subsubsection{Kingdomino}

Kingdomino is a game featuring \textit{enclosure}, \textit{open drafting},
and \textit{domino placement} mechanics for two, three, and four players.

The core gameplay consists of sequentially drafting four randomly
drawn domino-like tiles and placing them within a growing $5\times5$
grid, subject to adjacency constraints. A distinctive feature of Kingdomino
is its coupled action mechanism: each player selects one domino, which
directly determines that player’s position in the next turn order.
The more valuable a tile a player chooses, the worse their ranking
will be in the next round, and the fewer choices they will have at
that point. This introduces a trade-off between immediate reward (high-value
tiles) and strategic positioning (earlier versus later turn order),
creating a delayed and interdependent decision process. In addition,
scoring is non-linear, depending on the formation of connected domino
regions of the same landscape weighted by “crown” multipliers, found
on certain tiles. Rules of Kingdomino are available at boardgamegeek.com~\cite{bgg:kingdomino}.

One variant, called Mighty Duel, allows one to play on a $7\times7$
board. Both $5\times5$ and $7\times7$ variants are used as benchmarks
in the experiments presented in this paper.

In addition, the two variants studied use the following advanced rules:
``Middle Kingdom'' and ``Harmony'', which add additional end-game
scoring objectives related to tile placement.

It is NP-complete to decide whether a given score is achievable at
Kingdomino \cite{nguyen2020np}. An experimental study on multiplayer
search algorithms was conducted on Kingdomino (with four players)
\cite{gedda2018monte}.

Kingdomino notably won the Spiel des Jahres in 2017. According to
the BoardGameGeek community, Kingdomino ranks as the 80th-best family
game and 329th overall.

\subsubsection{Martian Dice}

Martian Dice offers a lightweight yet informative benchmark for evaluating
reinforcement learning and planning algorithms in stochastic games.
Although simpler than classical board games, it isolates core challenges
of decision-making under uncertainty: notably risk management and
more precisely the Push your luck dilemma. Each turn consists of repeated
roll-and-choose decisions, where an agent must determine whether to
stop and secure its current score or continue rolling to pursue higher
payoff at the risk of losing his recent gains.

In Martian Dice, players take on the role of Martians trying to abduct
humans, cows, and chickens from Earth. Each turn, a player rolls 13
custom dice showing different symbols: humans, cows, chickens, Earth’s
tanks, and death rays. After each roll, all tanks must be set aside.
The player then chooses one additional type to keep (one kind of Earthling
to abduct or death rays to defend against tanks), and rerolls the
remaining dice. This continues until all dice are set aside or decide
to stop. At the end of the turn, if the player has at least as many
death rays as tanks, the abductions are successful and the kept Earthlings
are scored. A bonus is awarded if the player manages to abduct at
least one human, one cow, and one chicken in the same turn. Complete
rules are available at boardgamegeek.com~\cite{bgg:martian_dice}.

The game is characterized by very limited player choices contrasted
with highly diverse random events, making it a particular case interesting
as benchmark for approximate decision-making methods.

According to the BoardGameGeek community, Martian Dice ranks as the
1,353rd-best game.

\subsubsection{Can’t Stop}

Can’t Stop is a game featuring push your luck, race, roll and move,
track movement mechanics.

In Can’t Stop, players roll four dice each turn and combine them into
pairs to advance markers up columns corresponding to dice sums. There
are 12 columns, each with a different difficulty. Players may iteratively
continue rolling to progress further or stop to secure their gains,
but risk losing all progress from the turn if no valid move is possible.
The goal is to be the first to complete three columns. Complete rules
are available at boardgamegeek.com~\cite{bgg:cantstop}.

According to the BoardGameGeek community, Can’t Stop ranks as the
217th-best family game and 765th overall.

\subsubsection{Shogun}

Shogun (1976) is played on an $8\times8$ board with pawns and a king,
aiming to capture the opponent’s pieces by moving onto their squares.
Each piece displays a number indicating how far it can move, and this
value changes randomly after each movement. Its complete rules are
available on Ludii~\cite{Piette2020Ludii}.

According to the BoardGameGeek community, Shogun has a complexity
comparable to Backgammon and ranks as the 1,074th-best family game
and 13,466th overall.
\end{document}